\documentclass[10pt]{article}
\usepackage{geometry}
\usepackage{cite}
\usepackage{amsmath}
\usepackage{graphicx}
\usepackage{appendix}
\usepackage{amsfonts}
\usepackage{subcaption}
\usepackage{textcomp}
\usepackage{gensymb}
\usepackage{lipsum}
\usepackage{indentfirst}
\usepackage[labelfont=bf]{caption}
\usepackage [english]{babel}
\usepackage [autostyle, english = american]{csquotes}
\MakeOuterQuote{"}

\usepackage{listings}
\usepackage{xcolor}

\definecolor{codegreen}{rgb}{0,0.6,0}
\definecolor{codegray}{rgb}{0.5,0.5,0.5}
\definecolor{codepurple}{rgb}{0.58,0,0.82}
\definecolor{backcolour}{rgb}{0.95,0.95,0.92}

\lstdefinestyle{mystyle}{
    backgroundcolor=\color{backcolour},   
    commentstyle=\color{codegreen},
    keywordstyle=\color{magenta},
    numberstyle=\tiny\color{codegray},
    stringstyle=\color{codepurple},
    basicstyle=\ttfamily\footnotesize,
    breakatwhitespace=false,         
    breaklines=true,                 
    captionpos=b,                    
    keepspaces=true,                 
    numbers=left,                    
    numbersep=5pt,                  
    showspaces=false,                
    showstringspaces=false,
    showtabs=false,                  
    tabsize=2
}

\title{\textbf{Modeling, Planning, and Control for Hybrid UAV Transition Maneuvers}}

\author{Spencer Folk}
\date{June 2020}

\usepackage[colorlinks=true,citecolor=blue,linkcolor=blue]{hyperref}

\usepackage{subfiles}

\begin{document}

\begin{titlepage}
   \begin{center}
       \vspace*{1cm}

       \huge
       \textbf{Modeling, Planning, and Control for Hybrid UAV Transition Maneuvers}

       \vspace{3cm}
       \Large
       Spencer Folk \\
       \large
       \vspace{0.2cm}
       sfolk@seas.upenn.edu
       
       \vspace{5.5cm}
       
       \large
       Qualifying Examination \\
       June 17$^{\text{th}}$, 2020 \\
       3-5 PM via Zoom
            
       \vspace{0.8cm}
       
       \large  
       Department of Mechanical Engineering and Applied Mechanics\\
       University of Pennsylvania\\
       Philadelphia, PA\\
       
       \vfill       
       \textbf{\underline{Committee}} \\
       \vspace{0.3cm}
       \large
       Dr. Cynthia Sung (Chair) \\
       Dr. Mark Yim (Advisor) \\
       Dr. Bruce Kothmann (Math Examiner)
            
   \end{center}
\end{titlepage}

\newpage

\begin{abstract}
    \textbf{Small unmanned aerial vehicles (UAVs) have become standard tools in reconnaissance and surveying for both civilian and defense applications. In the future, UAVs will likely play a pivotal role in autonomous package delivery, but current multi-rotor candidates suffer from poor energy efficiency leading to insufficient endurance and range. In order to reduce the power demands of package delivery UAVs while still maintaining necessary hovering capabilities, companies like Amazon are experimenting with hybrid Vertical Take-Off and Landing (VTOL) platforms. Tailsitter VTOLs offer a mechanically simple and cost-effective solution compared to other hybrid VTOL configurations, and while advances in hardware and microelectronics have optimized the tailsitter for package delivery, the software behind its operation has largely remained a critical barrier to industry adoption. Tailsitters currently lack a generic, computationally efficient method of control that can provide strong safety and robustness guarantees over the entire flight domain. Further, tailsitters lack a closed-form method of designing dynamically feasible transition maneuvers between hover and cruise. In this paper, we survey the modeling and control methods currently implemented on small-scale tailsitter UAVs, and attempt to leverage a nonlinear dynamic model to design physically realizable, continuous-pitch transition maneuvers at constant altitude. Primary results from this paper isolate potential barriers to constant-altitude transition, and a novel approach to bypassing these barriers is proposed. While initial results are unsuccessful at providing feasible transition, this work acts as a stepping stone for future efforts to design new transition maneuvers that are safe, robust, and computationally efficient. }
\end{abstract}

\hrulefill

{
  \hypersetup{linkcolor=black}
  \tableofcontents
}

\vspace{1cm}
\hrulefill

\newpage

\section{Introduction}
In the last decade, small unmanned aerial vehicles (UAVs) have been the subject of increased intellectual exploration in academic, industry, and defense settings. Applications for UAVs to date have ranged from Intelligence, Surveillance, and Reconnaissance (ISR) to civilian applications like bridge inspection, agriculture, and geological surveying. UAVs have even been employed to track and enforce social distancing of Italian citizens during the recent SARS-CoV-2 pandemic \cite{Holroyd:Coronavirus_Drones}. The future is bright for UAVs--stakeholders anticipate that small UAVs will soon be performing more difficult, transformative tasks such as autonomous delivery in large scale operations. As a token of the technology's potential, Amazon recently revealed a concept UAV with plans to launch a drone delivery service in suburban areas across the United States \cite{Vincent:Amazon}.

However, recent studies have exposed significant energy inefficiencies of traditional multi-rotor UAVs when operating at a large scale like in package delivery\cite{Zaremba:Drone_delivery}. In light of this, autonomous package delivery companies are experimenting with hybrid Vertical Take-Off and Landing (VTOL) aircraft, which ideally possess the hovering capabilities of a rotorcraft but the range and endurance comparable to fixed-wing airplanes. Hovering capabilities are important: they can potentially mitigate operational costs and logistical challenges by eliminating launch-and-recover infrastructure, such as the slingshots used by current life-saving drone delivery service, Zipline \cite{Jackson:Zipline}. Tailsitters are a variant of hybrid VTOLs that have reduced mechanical complexity compared to other UAV configurations (e.g. tilt-rotor or tilt-wing); unfortunately, there are many challenges hindering implementation of tailsitters for package delivery. Primarily, reduced mechanical complexity has meant relying on sophisticated controllers that can stabilize this underactuated system across all possible operating conditions. The design and evaluation of controllers for general tailsitters across all size scales remains a significant gap in the literature surrounding these aircraft.  

\begin{figure}[!ht]
    \centering
    \includegraphics[width=1\linewidth]{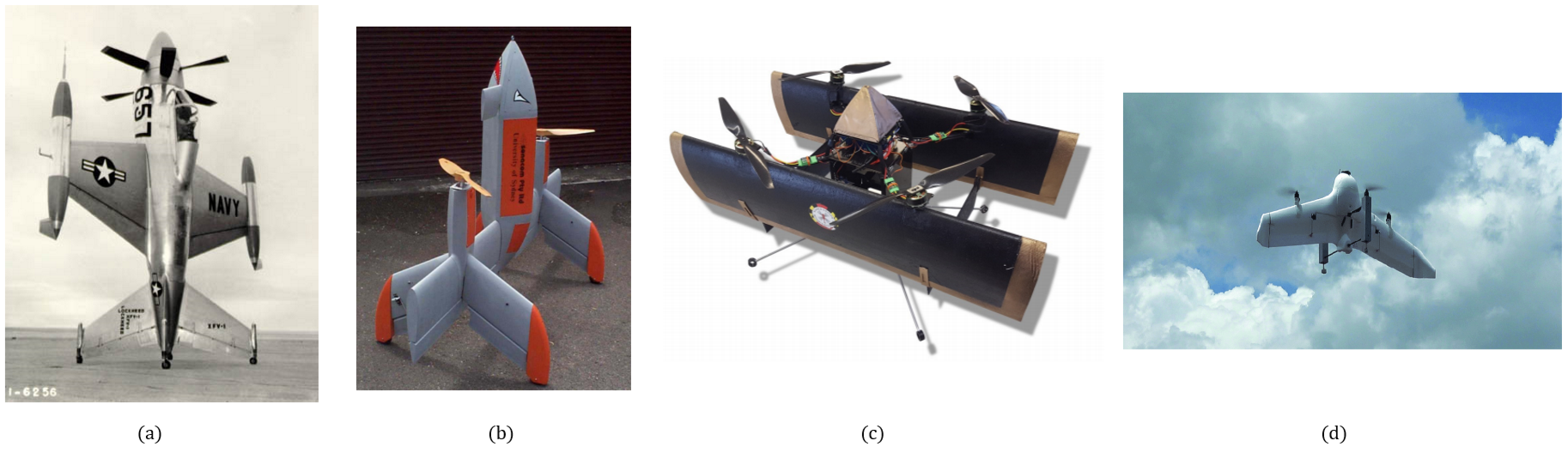}
    \caption{\small A brief selection of tailsitter aircraft showcasing the state of the art in hybrid UAV design over the years; \textbf{(a)} Lockheed XVF Pogo (1954); \textbf{(b)} Stone \textit{et. al} (2008,\cite{Stone:design}); \textbf{(c)} Phillips \textit{et. al.} (2017,\cite{Phillips:design}); \textbf{(d)} Gu \textit{et. al} (2019,\cite{Gu:design}).}
    \label{fig:aircraft}
\end{figure}

The first flying tailsitters were born out of Cold War era research and development of exotic aircraft. The definitive example of an operational tailsitter aircraft is the Lockheed XVF Pogo dating back to 1954. The Pogo was manual--take-off and landing required a skilled pilot--and only had 32 test flights before the project was shelved in 1955. Research on tailsitters was largely obscure for over half a century, until the miniaturization and affordability of aircraft components finally enabled tailsitter experimentation on a smaller scale and to a broader audience. Consequently, development of unmanned tailsitters for research purposes exploded with notable works by Stone \textit{et. al} in 2008, which were among the first to publish on the design and flight of small-scale T-wing tailsitter with consumer-grade electronics \cite{Stone:design}. Over the next few years, different research groups started publishing unique variants of the tailsitter; these include flying wing\cite{Sinha:design}\cite{Ritz:controller_nonlinear}, quad-wing\cite{Oosedo:design}, bi-plane\cite{Phillips:design}, and even Pogo-replica \cite{Knoebel:design} designs. Tailsitters have even been the object of studies regarding novel aircraft design methods, such as Gu \textit{et. al.} optimizing the tailsitter design via coordinate descent optimization\cite{Gu:design}. As impressive as these designs are, they have not surmounted the fundamental challenges facing tailsitters. 

While there have been great strides towards efficient and agile tailsitters, many of the challenges remaining pertain to the software behind these aircraft. One primary unsolved challenge is the need for generic and robust controllers that can safely stabilize tailsitters across their entire flight domain--this is difficult because the flight domain is quite large compared to traditional aircraft. Further, canonical methods of aircraft control have small stabilizing regions and are ill-equipped for the severely nonlinear aerodynamics in the post-stall regime. Approaches to controller design for tailsitters can be predominately classified as either linear or nonlinear. 

Linear control methods are tried and true, once acting as the backbone behind high-performance aircraft like fighter jets. However, as was the case with past-generation fighter jets, a linear controller for one tailsitter cannot be applied to another without extensive flight testing and tuning. Linear techniques rely on a discrete set of linearized models of the aircraft at different flight conditions. The simplest approach requires two controllers for linearized models at hover and forward flight, and relies on a pilot or open-loop maneuver to switch between these two modes\cite{Hochstenbach:design}. More sophisticated linear methods develop dozens or even hundreds of linearized models, each with a controller and corresponding gains, around the operating domain. For smooth operation, these approaches rely either on a high resolution between linearization points\cite{Reddinger:controller_linear_refframes}, or stitching sparse linearized models together through adaptive-model control\cite{Zhang:design} or gain-scheduling\cite{Silva:controller_linear}. More recently, Li \textit{et. al.} demonstrated Model Predictive Control (MPC) on a tailsitter linearized at hover, which could improve smooth switching between discrete models in the future\cite{Li:controller_MPC}. The common trait among these works is a nauseating amount of flight testing or simulation for a predetermined aircraft to improve controller performance. As with any linearization scheme, these controllers are unpredictable when operating far enough away from the nearest linearized model. This presents a safety and logistical concern, as the space of linearized models must cover the anticipated operating domain to ensure any sense of global stability.  

Researchers have also studied nonlinear approaches that in most cases require less meticulous flight testing, are agnostic to different tailsitter variants or scales, and generally provide broader stability guarantees. This literature can be further decomposed into coordinate transform methods\cite{Pucci:controller_nonlinear}, geometric representations of the dynamics\cite{Nogar:controller_nonlinear}, or optimal control strategies employing numerical analysis of the dynamics\cite{Zhou:nonlinear},\cite{Lyu:design}. Pucci \textit{et. al} proposes a clever change of coordinates that enables very simple controller design that stabilizes to a reference flight trajectory. However, this method leans on strong knowledge of the aerodynamics and an assumption of symmetry in the aircraft's body. In contrast to Pucci's approach, which is indifferent to size or even vehicle configuration provided the appropriate aerodynamics, Zhou \textit{et. al.} instead uses extensive wind tunnel testing akin to linear methods to build a high-fidelity model of their aircraft for a nonlinear controller. Geometric representations of a vehicle tracking a desired trajectory offer a middle-ground solution that defines the aircraft across a large operating domain, while still remaining lean and generic enough for applications to a broad set of tailsitters. One notable work by Ritz \textit{et. al.} combines an optimization routine with a geometric controller that performs online learning of the configuration's aerodynamics for global control\cite{Ritz:controller_nonlinear}. The literature thus far represents significant contributions towards global descriptions and control for tailsitters, but many of these nonlinear methods still rely on optimization schemes that increase the computational burden of the controller, and their extension to different scales has not been properly evaluated. 

A defining metric for tailsitter design and controller development is the transition maneuver, which moves the aircraft between hovering and forward flight. The transition maneuver is a good evaluation of a controller because it covers a large portion of the flight domain, and its difficulty has historically been a critical barrier for widespread adoption of tailsitter vehicles. More primitive transition maneuvers include the "stall-and-tumble" maneuver indicated by a large altitude gain (stall) followed by a drop (tumble) and glide into forward flight. This exercise in particular is often performed by manual pilots; in the autonomous case, it is an open-loop maneuver that can require up to a 20 meter drop depending on the size of the UAV. When considering package delivery over constrained airspaces like that in suburban or urban environments, the ideal transition maneuver would require little to no altitude change. Some approaches to accomplishing constant altitude transition formulate rudimentary trajectories and rely on the robustness of their controller to stabilize the tailsitter as best as possible\cite{Pucci:controller_nonlinear},\cite{Zhou:nonlinear}. In contrast, other researchers such as Oosedo \textit{et. al.} employ trajectory optimization to ensure transition is fast, dynamically feasible, and requires little control effort\cite{Oosedo:optimal}. Reddinger \textit{et. al.} contributes to this work by constraining transition maneuvers based on stall conditions and actuation limits\cite{Reddinger:controller_linear_refframes}. In all of these cases, trajectory optimization is performed offline due to computational burden which could prove to be another critical barrier to industry adoption. The ideal transition maneuver is efficient, can operate in the altitude-constrained airspaces of the future, and can be planned in real time on the vehicle's hardware.

Research and development of tailsitters thus far has produced impressive small-scale platforms that can operate over flight domains much larger than traditional multi-rotors or fixed-wing aircraft. Nonetheless, the lack of any comprehensive studies regarding the modeling, planning, and control of tailsitters in a scalable fashion represents an opportunity for intellectual and technological gains. In this paper, we derive a reduced-order dynamic model for thrust-actuated tailsitters and apply a nonlinear controller for stabilizing the vehicle to an arbitrary trajectory. Through a passive stability analysis of the dynamics, we isolate a phenomenon that makes transition at constant altitude very challenging, and we propose a novel method to bypass this phenomenon that leverages the dynamics of the aircraft. While this work does not relinquish reliance on a solid understanding of the vehicle's aerodynamics, it is an incremental step towards universal modeling, control, and planning for a scalable tailsitter aircraft. 

\newpage
\section{Methods and Theory}
With the goal of simulating dynamically feasible transition maneuvers for a hybrid UAV tailsitter, we now detail approaches to dynamic modeling, stability analysis, controller formulation, and trajectory generation for a tailsitter transition maneuver. We rely on a reduced-order model that is applicable to thrust-actuated tailsitters on a variety of scales.

\subsection{Vehicle Model}\label{sec:vehicle_model}

\begin{figure}[!ht]
    \centering
    \includegraphics[width=0.5\linewidth]{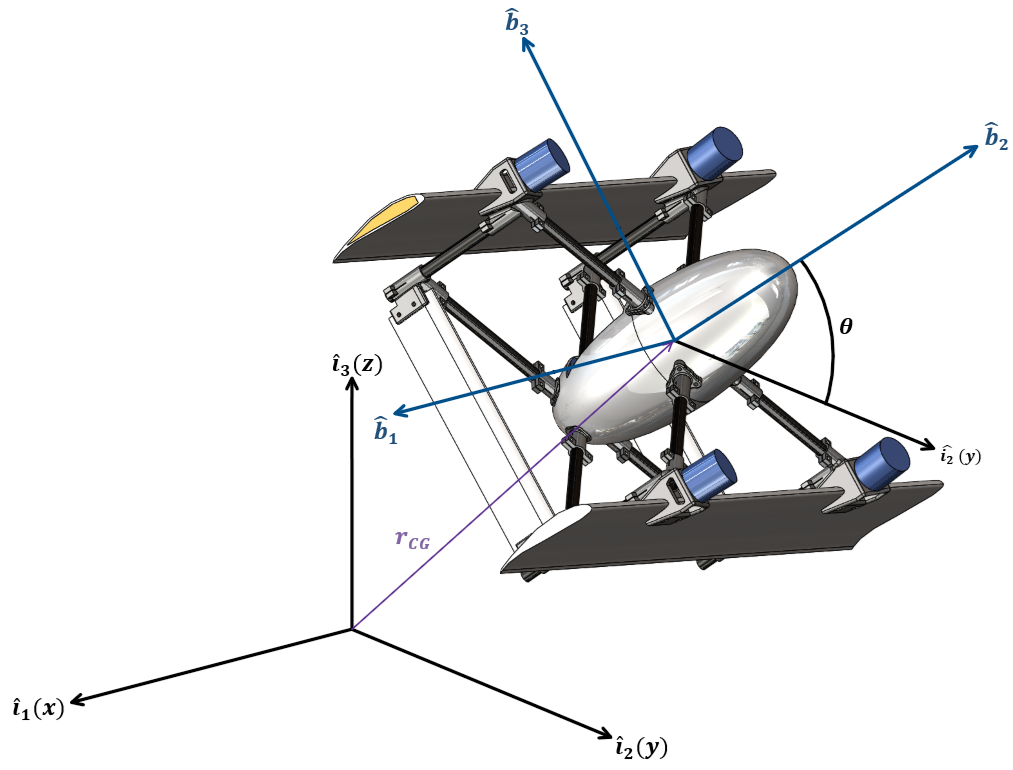}
    \caption{\small The Quadrotor Biplane Tailsitter (QBiT) configuration used as a motivating example for studying hybrid VTOL transition maneuvers. Attached to the QBiT is a body-fixed frame that is located in reference to a fixed inertial frame. For this project, only planar motion in the $\boldsymbol{\hat{i_2}}$-$\boldsymbol{\hat{i_3}}$ plane is considered, with changes only in the pitch axis by angle $\theta$. This CAD model was provided courtesy of Dr. Michael Avera from the United States Army Research Laboratory.}
    \label{fig:qbit_vehicle}
\end{figure}

The Quadrotor Biplane Tailsitter (QBiT) pictured in Figure \ref{fig:qbit_vehicle} is an ideal motivating example for the study of transition maneuvers for hybrid VTOL vehicles. The tailsitter features two parallel wings with two counter-rotating propellers on each wing (top and bottom) such that the overall configuration resembles a quadrotor. In the absence of control surfaces, the QBiT generates moments via differential thrust.  

The QBiT was first developed by the University of Maryland's Alfred Gessow Rotorcraft Center for the purpose of studying scalable unmanned aerial systems \cite{Phillips:design}. The design is intended to be produced at a variety of scales ranging from roughly $1$-kg to $20$-kg or more. However, the overall structure remains the same at all scales, enabling experimentation and evaluation of controllers across different vehicle sizes. The center module allows placement of fixed payloads--ideal for package delivery scenarios.

\subsection{Tailsitter Dynamics} \label{sec:tailsitter_dyn}

Below, we describe the governing equations for the dynamics of a thrust-actuated tailsitter in a planar side view. We consider only the planar dynamics because the transition maneuver is typically assumed to occur in these two dimensions\cite{Reddinger:controller_linear_refframes},\cite{Oosedo:optimal}. 

\subsubsection{Reference frames} \label{subsec:ref_frames}
\begin{figure}[!ht]
    \centering
    \includegraphics[width=1\linewidth]{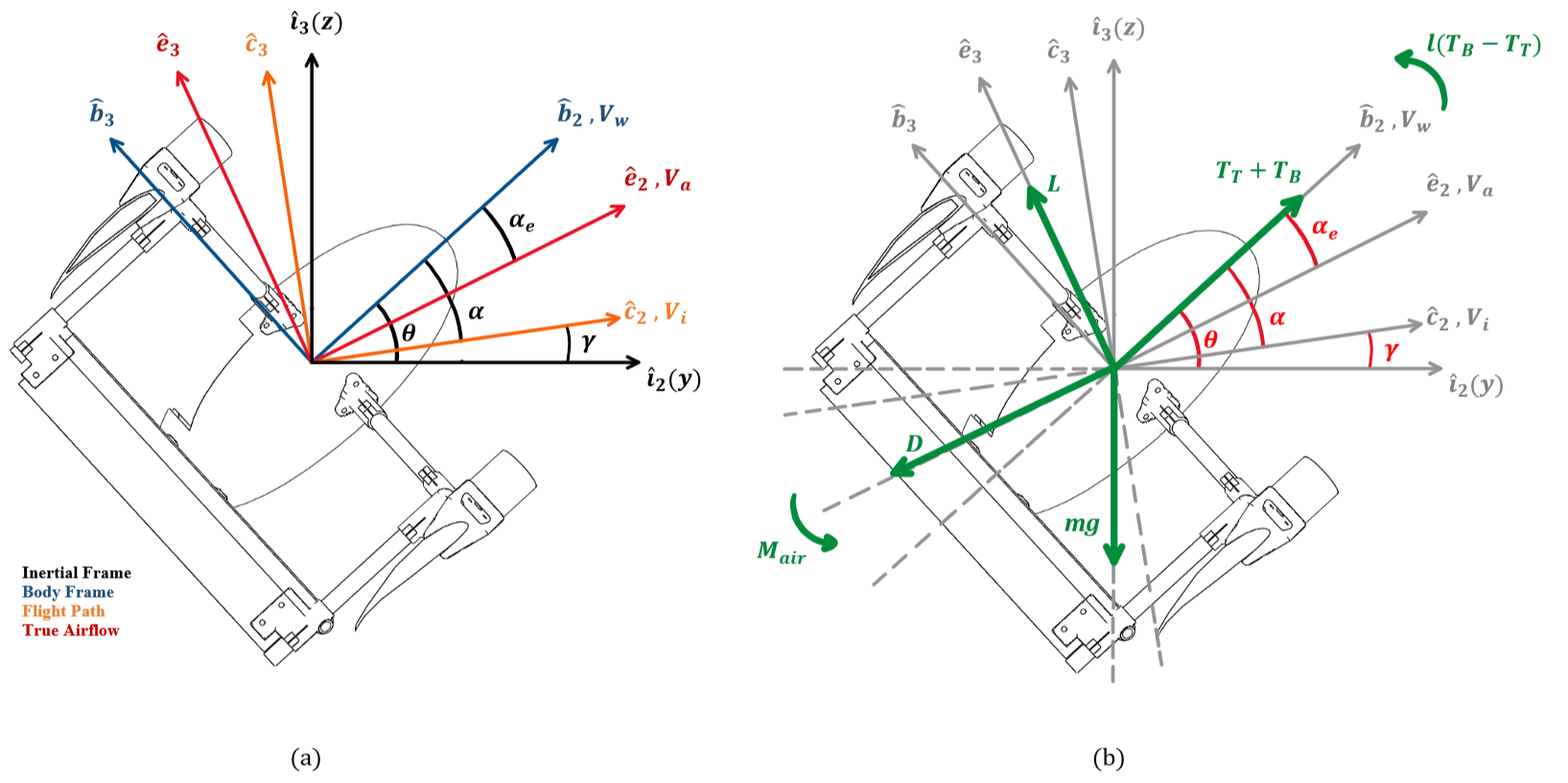}
    \caption{\small A side view, here defined in the y-z plane that shows \textbf{(a)} the reference frames $\mathcal{I}, \mathcal{B}, \mathcal{C}$ and $\mathcal{E}$ that are used to describe the vehicle's dynamics and orient different airflow contributions over the wing; \textbf{(b)} a free body diagram of the QBiT during transition flight.}
    \label{fig:ref_frame_fbd}
\end{figure}

There are four reference frames that are useful for modeling this unique hybrid aircraft: the inertial frame $\boldsymbol{\mathcal{I}} = \{\boldsymbol{\hat{i_1}, \hat{i_2}, \hat{i_3}}\}$, body frame $\boldsymbol{\mathcal{B}} = \{\boldsymbol{\hat{b_1}, \hat{b_2}, \hat{b_3} }\}$, flight path frame $\boldsymbol{\mathcal{C}} = \{\boldsymbol{ \hat{c_1}, \hat{c_2}, \hat{c_3} }\}$, and the true airflow frame  $\boldsymbol{\mathcal{E}} = \{\boldsymbol{ \hat{e_1}, \hat{e_2}, \hat{e_3}} \}$. These reference frames are illustrated in a planar view in Figure \ref{fig:ref_frame_fbd}\color{blue}a.\color{black} Note that the $\boldsymbol{\hat{i_1}, \hat{b_1}, \hat{c_1}},$ and $\boldsymbol{\hat{e_1}}$ axes point out of the page and the inertial frame is not fixed to the body.

The body frame $\boldsymbol{\mathcal{B}}$ is located on the vehicle's center of mass and is oriented such that $\boldsymbol{\hat{b_2}}$ is always parallel with the thrust plane of the rotors. The frames $\boldsymbol{\mathcal{C}}$ and $\boldsymbol{\mathcal{E}}$ are both located on a virtual aerodynamic center, which is fixed to the body regardless of the pressure distribution on the wings. These two frames are oriented based on different airflow velocities over the wings. Frame $\boldsymbol{\mathcal{C}}$ is aligned with the inertial velocity of the vehicle, $\boldsymbol{V_i}$, which represents the velocity of air across the wing due to vehicle translation. The magnitude of this velocity is related to the inertial $y$ and $z$ speeds as:
\begin{equation}
\begin{aligned}
    ||\boldsymbol{V_i}||^2 = \dot{y}^2 + \dot{z}^2
\end{aligned}
\label{eq:inertial_speed}
\end{equation}
and its orientation with respect to the horizon is defined by $\gamma := \arctan2(\dot{z},\dot{y})$.

Frame $\boldsymbol{\mathcal{E}}$ orients the aerodynamic forces by being aligned with the "true" airflow over the wing, $\boldsymbol{V_a}$, which is a vector sum of the inertial velocity and wake velocity in the inertial frame. The wake velocity, $\boldsymbol{V_w}$, describes the column of air moving across the wing due to the propeller down-wash, also referred to as "prop-wash". The wake velocity is oriented with the body frame $\boldsymbol{\hat{b_2}}$ axis, and its magnitude can be obtained via momentum theory \cite{stepniewski1984rotary}:
\begin{equation}
\begin{aligned}
    ||\boldsymbol{V_w}|| = \eta\sqrt{(||\boldsymbol{V_i}|| \cos \alpha)^2  + \frac{T}{\frac{1}{2}\rho \pi R^2}}
\end{aligned}
\label{eq:wind_speed}
\end{equation}
where $T$ is the thrust produced by a propeller, $\rho$ is the ambient air density, and $R$ is the radius of the propeller. Note the parameter $\eta \in [0,1]$: it is a propeller wake efficiency factor meant to reflect inefficiencies in the wake (e.g. turbulence or vortices) by discounting the contribution to the true airflow by the prop-wash. When $\eta = 0$, prop wash is ignored. In contrast, $\eta = 1$ represents fully ideal flow over the wing as calculated from momentum theory. This reduced-order model of the wake airflow had "good agreement" with blade element CFD simulations of a rotor-blown wing for speeds under approximately $8$-m/s \cite{Reddinger:controller_linear_refframes}. Above this speed, Reddinger \textit{et. al.} notes that the reduced-order model overpredicts the velocity of the air over the wing. 

The magnitude of $\boldsymbol{V_a}$ can be found by using the Law of Cosines on the triangle created by the vectors $\boldsymbol{V_i}, \boldsymbol{V_w}$, and $\boldsymbol{V_a}$. In other words, 
\begin{subequations}
\begin{align}
    ||\boldsymbol{V_a}|| = \sqrt{||\boldsymbol{V_w}||^2 + ||\boldsymbol{V_i}||^2 + 2||\boldsymbol{V_i}|| \text{ } ||\boldsymbol{V_w}|| \cos{\alpha}} \label{eq:true_airspeed_Va} \\
    \alpha := \theta - \gamma \label{eq:alpha}
\end{align}
\label{eq:true_airspeed}
\end{subequations}
where $\theta$ is the pitch angle of the aircraft. 

Because a significant portion of the wing is directly beneath the wake of the propellers, the airflow over the wing at low speeds can be dominated by this wake, as verified by Misiorowski \textit{et. al.} \cite{Misiorowski:rotor_wake_aero}. In this case, the actual angle of attack on the wing can be much lower than that estimated by $\alpha$ in Equation \ref{eq:alpha}. The effective angle of attack, $\alpha_e$, is the angle between the $\boldsymbol{\hat{b_2}}$ and $\boldsymbol{\hat{e_2}}$ axes. This angle is found by observing that $\boldsymbol{V_i}$ is the only contribution to $\boldsymbol{V_a}$ along the $-\boldsymbol{\hat{b_3}}$ axis.
\begin{equation}
\begin{aligned}
    \alpha_e = \arcsin{\frac{||\boldsymbol{V_i}|| \sin{\alpha}}{||\boldsymbol{V_a}||}}
\end{aligned}
\label{eq:effective_aoa}
\end{equation}
To provide some intuition for the airflow model described above, when prop-wash is ignored: $\eta, \boldsymbol{V_w} = 0 \implies \boldsymbol{V_a} = \boldsymbol{V_i} \implies \alpha_e = \alpha$. The reduced-order airflow model is a lean and generalizable representation of the important contributions from the propellers to the wing's airflow.

\subsubsection{Point-mass dynamic model} \label{subsec:point-mass_dynamic_model}
The free body diagram shown in Figure \ref{fig:ref_frame_fbd} enables the derivation of the vehicle dynamics for the QBiT. The model presented here describes the QBiT as a point-mass with gravity, aerodynamic forces, and motor thrusts acting on the body.

Restricting motion onto the y-z plane avoids any coupling terms in the rotational dynamics. The assumption that transition can occur in this plane is in agreement with other bodies of work in this area (see\cite{Reddinger:controller_linear_refframes},\cite{Pucci:controller_nonlinear}). The Cartesian dynamics in accordance with the free body diagram in Figure \ref{fig:ref_frame_fbd} can be written as the following system with states $\boldsymbol{x} = [y,z,\theta]^T$:
\begin{subequations}
\begin{align}
    &m\ddot{y} = (T_T + T_B)\cos{\theta} - L\sin{(\theta - \alpha_e)} - D \cos{(\theta-\alpha_e)} \label{eq:yddot}  \\
    &m\ddot{z} = -mg + (T_T + T_B) \sin{\theta} + L\cos{(\theta - \alpha_e)} - D \sin{(\theta - \alpha_e)} \label{eq:zddot} \\ 
    &I_{xx} \ddot{\theta} = M_{air} + l(T_B-T_T) \label{eq:thetaddot}
\end{align}
\label{eq:cartesian_dynamics}
\end{subequations}
In this system, $m$ indicates the vehicle mass, $g$ is the gravitational constant, $I_{xx}$ is the principal moment of inertia about the $\boldsymbol{\hat{i_1}}$ (x) axis, and $l$ denotes the distance (as measured along $\boldsymbol{\hat{b_3}}$ axis) between each wing. The variables $T_T$ and $T_B$ are the thrust values of the top and bottom \textit{sets} of propellers. The variables $L$ and $D$ are lift and drag forces, which are defined to be functions of angle of attack and airflow over the wing, i.e. $L = L(\alpha_e, \boldsymbol{V_a})$ and $D = D(\alpha_e, \boldsymbol{V_a})$. $M_{air}$ denotes any pitching moments created by the aerodynamic forces across the wing, similarly $M_{air} = M_{air}(\alpha_e, \boldsymbol{V_a})$. 

System \ref{eq:cartesian_dynamics} was purposefully chosen to strike a balance between a simple representation of a planar tailsitter and a higher-fidelity model that might necessitate the use of computational methods for analysis. 

\subsubsection{Underlying assumptions} \label{subsec:assumptions}
The dynamics have been derived based on some key underlying assumptions. We assume that the lift and drag forces can be approximated as acting on an aerodynamic center for each wing, which can then be averaged into a single virtual point along the $\boldsymbol{\hat{b_2}}$ axis by taking advantage of the aircraft's symmetry. We also assume that this virtual aerodynamic center is coincident with the vehicle's center of gravity to further simplify the moment equation. This is possible on smaller platforms with thoughtful placement of the battery, electronics, and payload in the center module. For the thrust inputs, we assume that instantaneous changes in thrust can be achieved. Lastly, we assume that motion occurs only in the y-z plane by ensuring both motors on each wing are synchronized, which is to say that no incidental roll ($\boldsymbol{\hat{b_2}}$) or yaw ($\boldsymbol{\hat{b_3}}$) is generated by the propellers. 
\subsection{Aerodynamics} \label{sec:aero}
\begin{figure}
\makebox[0.95\linewidth][c]{
    \centering
\begin{subfigure}{0.35\textwidth}
  \centering
  \includegraphics[width=1.1\linewidth]{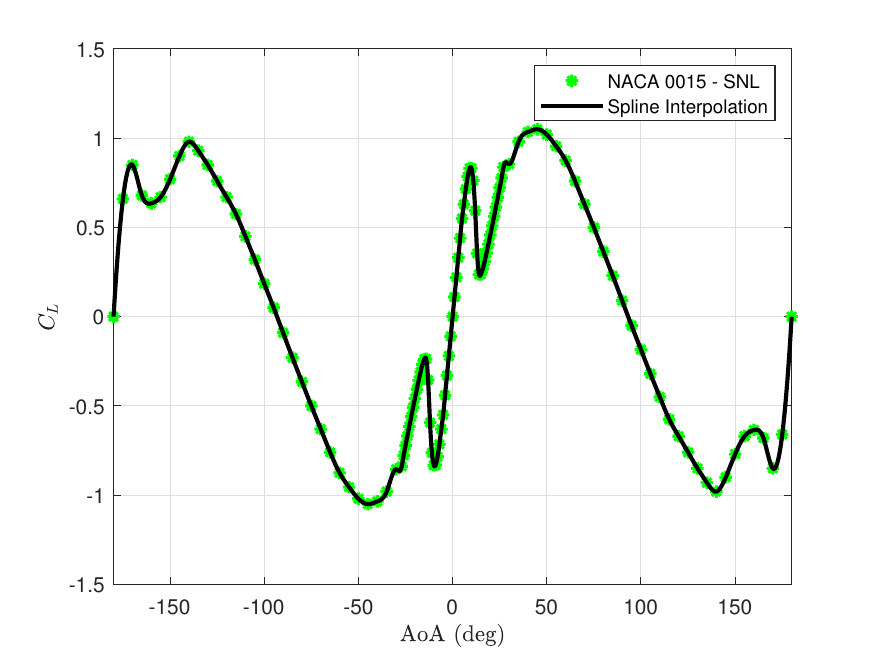}
  \caption{}
  \label{fig:aero_cl}
\end{subfigure}
\begin{subfigure}{.35\textwidth}
  \centering
  \includegraphics[width=1.1\linewidth]{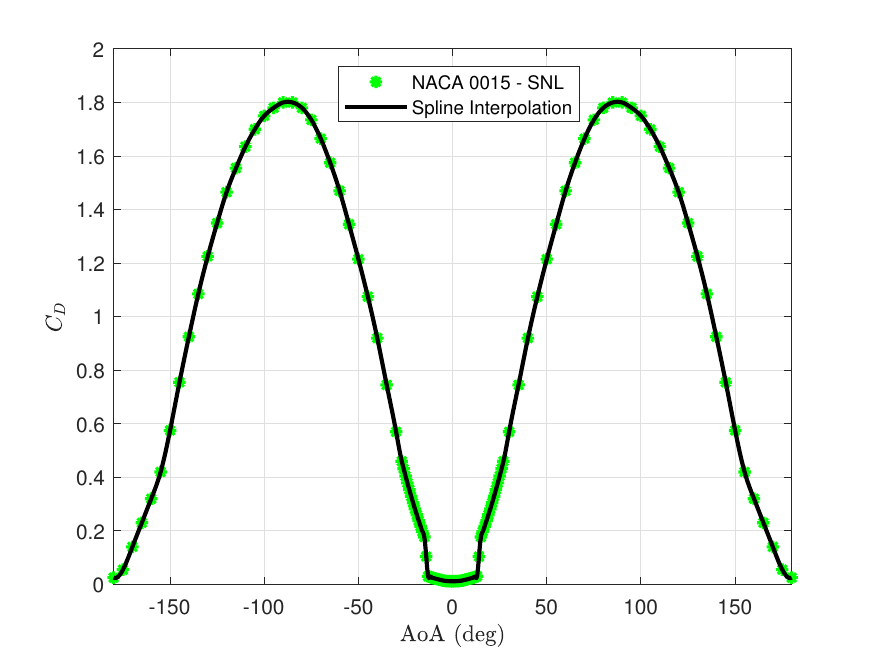}
  \caption{}
  \label{fig:aero_cd}
\end{subfigure}
\begin{subfigure}{.35\textwidth}
    \centering
    \includegraphics[width=1.1\linewidth]{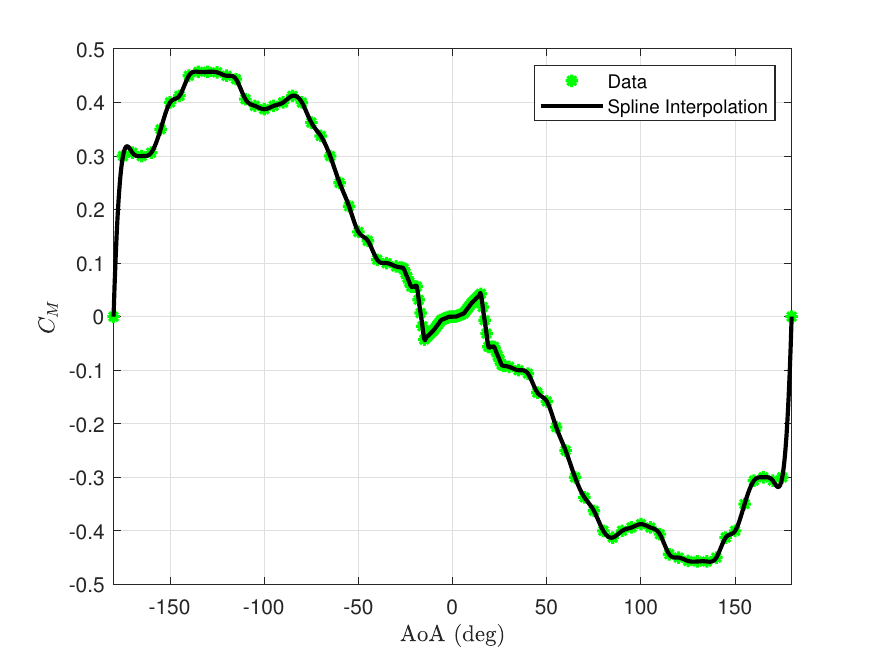}
    \caption{}
    \label{fig:aero_cm}
\end{subfigure}
}
\caption{\small \textbf{(a)} Lift, \textbf{(b)} Drag, and \textbf{(c)} Pitching Moment aerodynamic coefficients from -180$\degree$ to 180$\degree$ for a symmetric NACA 0015 airfoil at $Re = 160,000$. Data was taken from Sandia National Laboratories wind tunnel test data\cite{Sheldahl:aero_data} and fitted with a cubic spline to maintain smoothness in the aerodynamics.}
\label{fig:aero_fns}
\end{figure}

Aerodynamics play an important role in the dynamics of the QBiT through $L, D, $ and $M_{air}$. In the quasi-steady-state model shown here, the coefficients of lift ($C_L$), drag ($C_D$), and pitching moment ($C_M$) are unique to the airfoil on the vehicle; given that hybrid vehicles such as the QBiT operate over such a large flight domain, these coefficients must be defined for an unusually large range of angle of attack. Wind tunnel data was taken for symmetric airfoils from Sandia National Laboratories\cite{Sheldahl:aero_data}, which covers the full 360$\degree$ pitch that the QBiT might experience during flight. In Figure \ref{fig:aero_fns}, wind tunnel tests for a NACA 0015 symmetric airfoil at $Re = 160,000$ (corresponding to airspeeds on the order of $10$-m/s) are presented for $\alpha \in [-180, 180]$, fitted with a cubic spline interpolation. We assume that the center module contributions to the aerodynamics are negligible.

The quasi-steady-state approximation of the forces and moments generated by the wing reduces the aerodynamics to a dependence only on airspeed and angle of attack, at the cost of neglecting more nuanced--yet appreciable--behavior such as dynamic stall. Hence, the aerodynamic lift, drag, and pitching moment on the wing are found by
\begin{subequations}
\begin{align}
    L = \frac{1}{2}\rho ||\boldsymbol{V_a}||^2 S_{wing} C_L(\alpha_e) \label{eq:lift_force} \\ 
    D = \frac{1}{2}\rho ||\boldsymbol{V_a}||^2 S_{wing} C_D(\alpha_e) \label{eq:drag_force} \\
    M_{air} = \frac{1}{2} \bar{c} \rho ||\boldsymbol{V_a}||^2 S_{wing}  C_M(\alpha_e) \label{eq:pitch_moment}
\end{align}
\label{eq:aero_forces}
\end{subequations}
where $\bar{c}$ is the chord of the wing and $S_{wing}$ represents the planform area of the wing ($S_{wing} = \bar{c}b$). The aerodynamic coefficients--$C_L(\cdot), C_D(\cdot),$ and $C_M(\cdot)$--are defined by the cubic spline interpolations as seen in Figure \ref{fig:aero_fns}. This is done to preserve smoothness in the aerodynamics to work well with the controller. Some notable properties of symmetric airfoils are: $C_D(-\alpha) = C_D(\alpha)$ and $C_L(-\alpha) = -C_L(\alpha)$. 

Typically wind tunnel measurements can be very noisy in the post-stall region ($\alpha>15\degree$) due to turbulent and chaotic effects; for this reason, most airplanes are designed to operate within the stall limit, and those that do operate in the post-stall regime undergo exhaustive instrumented flight tests to validate aerodynamic models. The aerodynamic model presented here is commonly used for controller synthesis and low-fidelity simulation\cite{Ritz:controller_nonlinear}\cite{Reddinger:controller_linear_refframes}. 
\subsection{Passive Stability Analysis} \label{sec:stability_analysis}

Stability analysis gives insight into an aircraft's ability to track a desired flight path. Basic stability analysis, such as that introduced by Etkin, often measures the \textit{pitch stiffness} associated with the aircraft. Pitch stiffness is the approximate slope of the pitching moment coefficient as a function of the angle of attack. For a flying wing configuration such as the QBiT, passive stability at an equilibrium angle of attack requires both that the pitching moment is zero and its slope is negative. In other words, a small increase in angle of attack from equilibrium would produce a "nose-down" (negative) pitching moment to restore the aircraft to the equilibrium \cite{Etkin:stability_analysis}. 

For typical aircraft with pitch control surfaces, e.g. elevators or elevons, the pitch stiffness of the aircraft can be altered by control inputs to stabilize different angles of attack or produce desirable responses to disturbances. On the contrary, pitching moments on the QBiT can only be produced by differential thrust. This motivates Pucci's approach to assessing the stability to a desired flight path\cite{Pucci:controller_nonlinear}. This method identifies passive equilibrium angles of attack from the nonlinear dynamics, and then determines the stability of those equilibria by linearizing the system at a nominal flight velocity. 

Pucci's stability analysis begins by forming an equation for equilibrium angles of attack, first by considering the forces acting on a point-thrust VTOL vehicle--those presented in Figure \ref{fig:ref_frame_fbd}b--projected onto the body frame $\boldsymbol{\mathcal{B}}$ in steady state (trim) flight. 

\begin{subequations}
\begin{align}
    &\boldsymbol{\hat{b_2}}: ma_{b2} = L \sin \alpha_e - D \cos \alpha_e - mg \sin \alpha + (T_T + T_B)  \label{eq:stability_dynamics_b2} \\
    &\boldsymbol{\hat{b_3}}: ma_{b3} = L \cos \alpha_e + D \sin \alpha_e - mg \cos \alpha \label{eq:stability_dynamics_b3}
\end{align}
\label{eq:stability_dynamics}
\end{subequations}
where $a_{b2}, a_{b3}$ are placeholder values for the body accelerations. Here we are interested in Equation \ref{eq:stability_dynamics_b3} for assessing passive stability because the control inputs, $T_T + T_B$, do not influence the $\boldsymbol{\hat{b_3}}$ axis. We assume steady state flight ($a_{b2}, a_{b3} = 0$) and arrange Equation \ref{eq:stability_dynamics_b3} into the following form: 
\begin{subequations}
\begin{align}
    & 0 = \cos \alpha - a_v C_L(\alpha_e) \cos \alpha_e - a_v C_D(\alpha_e) \sin \alpha_e \label{eq:stability_dynamics_aoa} \\ 
    &a_v := \frac{ \frac{1}{2} \rho S_{wing} ||\boldsymbol{V_a}||^2}{mg} \label{eq:a_v}
\end{align}
\label{eq:stability_dynamics_a_v}
\end{subequations}
where $a_v$ is a dimensionless variable that we will refer to as \textit{aerodynamic loading}. A necessary condition for stabilization to a desired airspeed is the existence of a pair $\{\alpha, \alpha_e\}$ that satisfies Equation \ref{eq:stability_dynamics_aoa}. Equation \ref{eq:stability_dynamics_b2} produces a second condition, but it can be satisfied by an arbitrary selection of $(T_T + T_B)$ once $\{\alpha, \alpha_e\}$ is determined to satisfy Equation \ref{eq:stability_dynamics_aoa}. Here, we have reduced stability to general conditions on the tailsitter's aerodynamics. 

For the remainder of this stability analysis, we will assume the propeller downwash can be ignored: $\eta, \boldsymbol{V_w} = 0 \implies \alpha_e = \alpha$. However, note that if we included propeller downwash, we would see control inputs enter Equation \ref{eq:stability_dynamics_aoa} suggesting the possibility for any arbitrary angle of attack to be stabilized. This will be left for future work. We now solve Equation \ref{eq:stability_dynamics_aoa} for the aerodynamic loading.
\begin{equation}
    a_v = \frac{\cot \alpha}{C_D(\alpha) + C_L(\alpha) \cot \alpha}
\label{eq:a_v_aoa}
\end{equation}

\begin{figure}[!ht]
    \centering
    \begin{subfigure}{0.4\textwidth}
        \centering
        \includegraphics[width=1\linewidth]{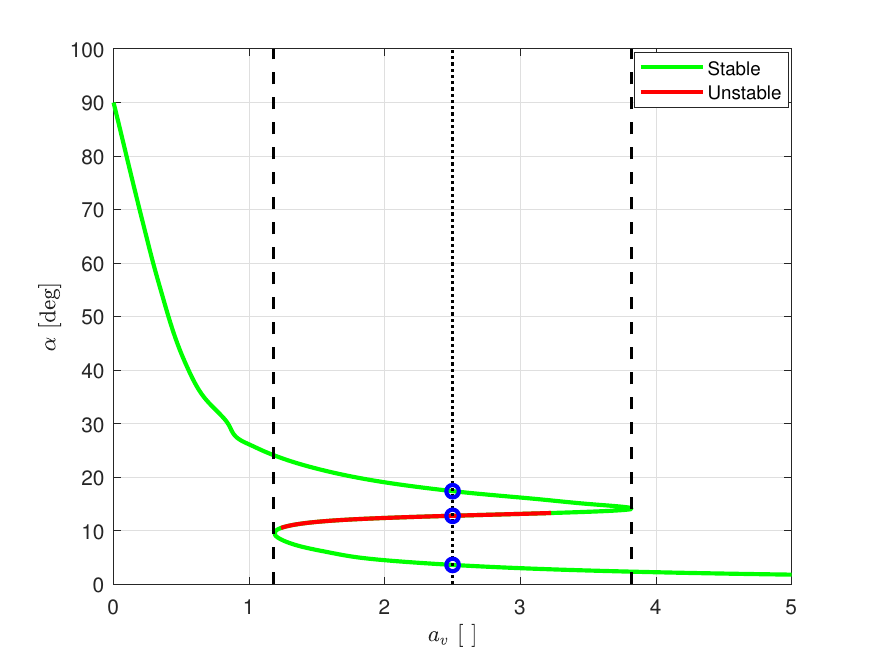}
        \caption{}
        \label{fig:bifurcation_aoa}
    \end{subfigure}
    \begin{subfigure}{0.4\textwidth}
        \centering
        \includegraphics[width=1\linewidth]{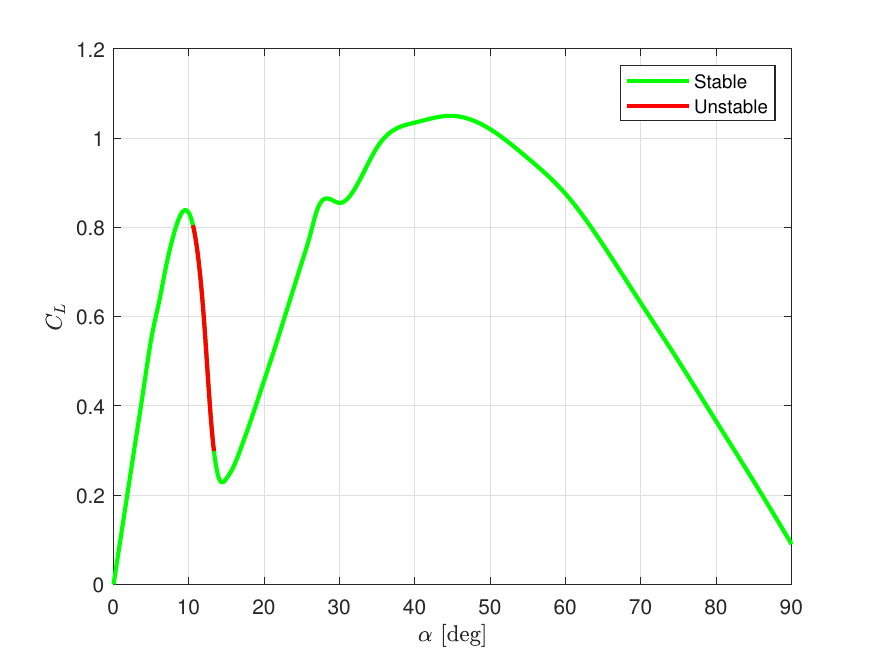}
        \caption{}
        \label{fig:bifurcation_lift}
    \end{subfigure}
    \caption{\small \textbf{(a)} Equilibria angles of attack associated with a desired airspeed characterized by the dimensionless parameter $a_v$. The equilibria are colored by stability criterion described in Equation \ref{eq:routh_hurwitz_conditions} derived by Pucci \cite{Pucci:controller_nonlinear}. The bifurcation region is bounded by dashed lines, and a sample of three equilibria $\alpha =\{3.63\degree, 12.8\degree, 17.4\degree\}$ are selected for $a_v = 2.5$; \textbf{(b)} The unstable equilibria are observed to coincide with the stall region associated with a NACA 0015 symmetric airfoil.}
    \label{fig:bifurcation_plots}
\end{figure}

Provided functions or approximations for the aerodynamic coefficients, $C_L(\cdot)$ and $C_D(\cdot)$, we can now assess the \textit{existence} of equilibrium angles of attack. Recall from Section \ref{subsec:ref_frames} that in planar motion, $\alpha = \theta - \gamma$, so given a desired flight path angle $\gamma$ and the equilibria $\alpha$, we can solve for the equilibrium body orientation of the vehicle. The equilibria are plotted in Figure \ref{fig:bifurcation_aoa} for a NACA 0015 symmetric airfoil with lift and drag coefficients described by Figure \ref{fig:aero_fns}.

As seen in Figure \ref{fig:bifurcation_aoa}, the structure of Equation \ref{eq:a_v_aoa} leads to a bifurcation of possible solutions, which is a common observation in classical studies of nonlinear systems and control. Often times these bifurcations occur due to a change in control input, such as a constant torque applied to an inverted pendulum resulting in a pitchfork bifurcation in the stable pendulum angle. Extending this analogy, we can think of the aerodynamic loading, $a_v$, as a "virtual control input". As we vary the airspeed by changing $a_v$, the equilibria solutions shift, appear, and disappear! In Figure \ref{fig:bifurcation_aoa}, for example, another solution forms at $a_v = 1.18$ and forks into two equilibria immediately. At $a_v = 3.82$, two equilibria meet and annihilate one another. Notably, three equilibrium orientations exist in the region $a_v \in (1.18, 3.82)$.

The next task is determining the stability characteristics of these equilibrium solutions, which is fully described in the work of Pucci (see\cite{Pucci:thesis}, Appendix A.8 p. 143). Fundamentally, an equilibrium angle of attack is determined to be stable if the real parts of the eigenvalues corresponding to the linearized error dynamics from System \ref{eq:cartesian_dynamics} are not positive. The aforementioned error dynamics, linearized around a desired inertial velocity $\boldsymbol{V_{ref}} = \dot{y}_r \boldsymbol{\hat{i_2}} + \dot{z}_r \boldsymbol{\hat{i_3}}$ are: 
\begin{equation}
    m\ddot{\boldsymbol{e}} \approx \frac{1}{2} \rho S_{wing} \frac{\partial}{\partial \boldsymbol{\dot{\boldsymbol{r}}}} \Bigr \{ |\boldsymbol{\dot{r}} | \begin{bmatrix} -C_D(\alpha) & -C_L(\alpha) \\ C_L(\alpha) & -C_D(\alpha) \end{bmatrix} \boldsymbol{\dot{r}} \Bigr \} _{\boldsymbol{\dot{r}} = \boldsymbol{V_{ref}}} \dot{\boldsymbol{e}}
\label{eq:pucci_stability}
\end{equation}
where $\boldsymbol{r} = [y, z]^T$ is the position of the body in the inertial frame and $\boldsymbol{e}$ represents the state error. 

Notice that the argument of $\frac{\partial}{\partial \boldsymbol{V_i}} \{ \cdot \}$ is the vector expression of the aerodynamics, Equation \ref{eq:aero_forces}, written in the inertial frame. The evaluation of this partial derivative is too lengthy to include in the main body of this paper, but the signs of the eigenvalues for this partial derivative can be inferred from the eigenvalues of the following matrix: 
\begin{equation}
    \begin{bmatrix} -2C_D(\alpha) & C'_D(\alpha) - C_L(\alpha) \\ 2C_L(\alpha) & -C'_L(\alpha) - C_D(\alpha) \end{bmatrix}
\label{eq:stability_eigs}
\end{equation}
where the superscript $(\cdot)'$ denotes the derivative with respect to the angle of attack, $\alpha$. The characteristic polynomial for this matrix is: 
\begin{subequations}
\begin{align}
    &\lambda^2 + p(\alpha) \lambda + 2q(\alpha) = 0 \label{eq:stability_char_poly} \\
    &p(\alpha) := 3C_D + C'_L \label{eq:p_stability} \\
    &q(\alpha) := C_D^2 + C_DC'_L - C_LC'_D + C_L^2 \label{eq:q_stability}
\end{align}
\label{eq:stability_criterion}
\end{subequations}
Finally, we can apply Routh-Hurwitz stability criterion for a second order polynomial to determine when Equation \ref{eq:stability_criterion} has negative real solutions, which will signify exponential growth in the system's response. Here we arrive at the final conditions for which an equilibrium angle of attack $\alpha$ satisfying Equation \ref{eq:stability_dynamics_aoa} is unstable: 
\begin{equation}
\begin{aligned}
    p(\alpha)q(\alpha) < 0 \text{\space \space \space \textbf{OR}  \space \space \space   } p(\alpha) < 0 \text{  and  } q(\alpha) < 0
\end{aligned}
\label{eq:routh_hurwitz_conditions}
\end{equation}

In Figure \ref{fig:bifurcation_plots}, we apply the conditions above to a NACA 0015 airfoil to arrive at an estimate of the stability for an equilibrium given as a solution to Equation \ref{eq:stability_dynamics_aoa}. The stability criterion is formulated on a linearization of the dynamics presented in System \ref{eq:cartesian_dynamics}, and as such the unstable region is an estimate. However, classical understandings of bifurcation leads to the conclusion that this unstable region extends from $a_v \in (1.18, 3.82)$. The existence of multiple equilibria for a given airspeed makes transitioning between hover ($\alpha = 90\degree$) and forward flight ($\alpha \leq 15\degree$) non-trivial. This will be further investigated in the results section. 
\subsection{The Nonlinear Geometric Controller} \label{sec:geometric_controller}

\begin{figure}[!ht]
    \centering
    \includegraphics[width=1\linewidth]{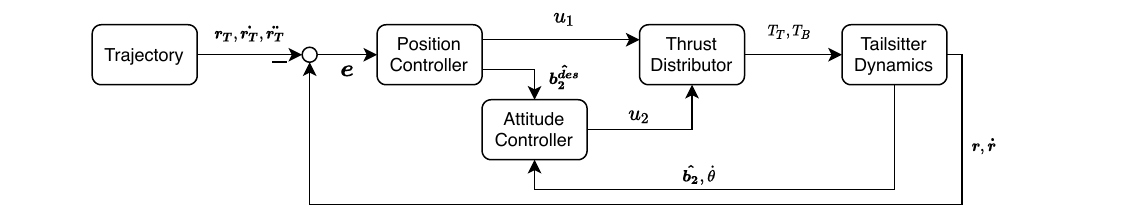}
    \caption{The control block diagram describing the nonlinear geometric controller implemented for stabilization to a desired trajectory. For clarity, the position controller houses Equations \ref{eq:pos_controller}--\ref{eq:u1}, the attitude controller Equations \ref{eq:b2_des}--\ref{eq:u2}, and the thrust distributor Equation \ref{eq:thrust_commands}.}
    \label{fig:control_block_diagram}
\end{figure}

Armed with the dynamics presented in Section \ref{sec:tailsitter_dyn}, we will now present a controller to stabilize 2-D trajectories. We enhance a nonlinear geometric controller--formalized and demonstrated for quadrotors by Lee \textit{et. al.}\cite{Lee:controller_nonlinear_geometric} and Mellinger \textit{et. al.}\cite{Mellinger:controller_nonlinear_geometric}, respectively--to handle aerodynamic forces and moments. Outputs from a position controller feed into an attitude controller in series leading to a cascaded, hierarchical control policy illustrated in Figure \ref{fig:control_block_diagram}. The control policy is "geometric" because it is constructed on a geometric representation of the thrust acceleration vectors in $SE(3)$.

The goal is to stabilize the position of the tailsitter, $\boldsymbol{r} = [y,z]^T$, along a \textit{known trajectory} characterized by $\boldsymbol{z_T} = [\boldsymbol{r_T}, \boldsymbol{\dot{r_T}}, \boldsymbol{\ddot{r_T}}]^T$. We can formulate this in the form of second-order error dynamics, $\boldsymbol{e} = \boldsymbol{r - r_T}$. 
\begin{equation}
    \boldsymbol{\ddot{e}} + 2\zeta\omega_n\boldsymbol{\dot{e}} + \omega_n^2\boldsymbol{e} = 0
    \label{eq:PD_error}
\end{equation}

To produce error dynamics defined by a desirable damping, $\zeta$, and natural frequency, $\omega_n$, we select $K_p = \omega_n^2$ and $K_d = 2\zeta\omega_n$. For this study, values for $K_p$ and $K_d$ are selected based on critical damping ($\zeta = 1$) and an $\omega_n$ that satisfies a desired settling time. Simulation gains and their associated units are presented in the Appendix, Table \ref{table:parameters}. 

The position controller begins by computing the desired acceleration by rearranging Equation \ref{eq:PD_error} and measuring the difference between the current state and desired trajectory.  
\begin{equation}
    \ddot{\textbf{r}}_{des} = \ddot{\boldsymbol{r}_T} - \boldsymbol{K_d}(\dot{\boldsymbol{r}} - \dot{\boldsymbol{r}_T}) - \boldsymbol{K_p}(\boldsymbol{r} - \boldsymbol{r}_T) 
    \label{eq:pos_controller}
\end{equation}
Notice here that $\boldsymbol{K_p}$ and $\boldsymbol{K_d}$ are now $2\times2$ diagonal gain matrices, and $\ddot{\boldsymbol{r}_T}$ acts as a feedforward acceleration term.

Once the desired acceleration is computed from Equation \ref{eq:pos_controller}, we use System \ref{eq:cartesian_dynamics} in Section \ref{subsec:point-mass_dynamic_model} to define a desired thrust vector, $\boldsymbol{F^{des}}$. The desired thrust vector is derived from solving the translational dynamics (Equations \ref{eq:yddot}, \ref{eq:zddot} in vector form) for the control input $u_1 = T_T + T_B$ and substituting the desired acceleration. 
\begin{equation}
    \boldsymbol{F^{des}} = m\ddot{\boldsymbol{r}}_{des} + \begin{bmatrix} 0 \\ mg \end{bmatrix} - [\boldsymbol{^IR_E]} \begin{bmatrix} -D \\ L \end{bmatrix}
\label{eq:F_des}
\end{equation}
where the matrix $\boldsymbol{[^IR_E]}$ is a $2\times2$ rotation matrix representing a counter-clockwise rotation by the angle $\theta - \alpha_e$. The desired force enables explicit computation of $u_1$ by projecting $\boldsymbol{F^{des}}$ onto the $\boldsymbol{\hat{b_2}}$ axis of the body frame expressed in the inertial frame:
\begin{equation}
u_1 = \boldsymbol{\hat{b_2}}^T \boldsymbol{F^{des}}
\label{eq:u1}
\end{equation}

Up until this point, we have proposed a method to match the total thrust of the tailsitter, $u_1$, with the magnitude of the desired force $||\boldsymbol{F^{des}}||$ along the vehicle's controllable axis, $\boldsymbol{\hat{b_2}}$. The task now is to orient the body frame axis with the direction of the desired force as measured in the inertial frame. This can be accomplished via attitude control to a desired body orientation, determined by the orientation of $\boldsymbol{\hat{b_2}}$ in the inertial frame. We can define the desired body orientation from $\boldsymbol{F^{des}}$: 
\begin{equation}
\boldsymbol{\hat{b_2}^{des}} = \frac{\boldsymbol{F^{des}}}{|| \boldsymbol{F^{des}} ||} \\
\label{eq:b2_des}
\end{equation}

We calculate the attitude error, $e_\theta$, simply by determining the dot product between the current body orientation $\boldsymbol{\hat{b_2}}$, and the desired $\boldsymbol{\hat{b_2}^{des}}$, and feed that error into the following attitude control law to determine the desired moment, or $u_2$.
\begin{subequations}
\begin{align}
    u_2 = I_{xx}(-K_R e_\theta - K_\omega \dot{\theta})  - M_{air} \\
    e_\theta := \angle(\boldsymbol{b_2}, \boldsymbol{b_2^{des}})
\end{align}
\label{eq:u2}
\end{subequations}
where $K_R$ and $K_\omega$ are proportional and derivative gains associated with the pitch axis. Note that the attitute rate along the transition trajectory is assumed to be small or zero, so it is sufficient to say that $\dot{e_\theta} \approx \dot{\theta} $. This expression for $u_2$ was derived in a similar fashion to $u_1$ by solving for the control moment $u_2 = l(T_B-T_T)$ in Equation \ref{eq:thetaddot} and replacing $\ddot{\theta}$ by a desired angular acceleration defined by second order error dynamics, as in Equation \ref{eq:PD_error}. 

We can now solve for the thrust values of each set of motors ($T_T$ and $T_B$) by solving the following system of equations, which notably is identical to that of a planar quadrotor formulation: 

\begin{equation}
\begin{bmatrix} 1 & 1 \\ -l & l \end{bmatrix} \begin{bmatrix} T_T \\ T_B \end{bmatrix} = \begin{bmatrix} u_1 \\ u_2 \end{bmatrix}
\label{eq:thrust_commands}
\end{equation}
where $l$ represents the length between the rotors as measured along the $\boldsymbol{\hat{b_3}}$ axis. Notice that the "$A$" matrix in this linear equation is always full rank unless $l = 0$. Therefore a solution \{$T_T, T_B$\} will always exist for nonzero inputs \{$u_1, u_2$\}.
\subsection{Trajectory Generation} \label{sec:traj_gen}

In Section \ref{sec:geometric_controller}, we formulate a controller that stabilizes the QBiT along an arbitrary known trajectory. In this section we describe two methods of trajectory generation that transition the vehicle from a hover to forward flight. In both cases, we are interested in a trajectory that transitions the vehicle at a constant altitude to be better suited for the constrained airspaces of the future. Accordingly, trajectories are formulated along the $\boldsymbol{\hat{i_2}} (y)$ axis only. 

\subsubsection{Constant acceleration} \label{subsec:traj_const_acc}

The simplest transition maneuver is, naively, one which requests a constant horizontal acceleration until a desired cruising speed, denoted $V_s$, is achieved. The formulation of this trajectory is rather straightforward: it is characterized only by a desired forward acceleration, $a_s$. The transition is defined by:  
\begin{subequations}
\begin{align}
    \ddot{y_r}(t) = a_s \\ 
    \dot{y_r}(t) = V_0 + a_s t \\ 
    y_r(t) = \frac{1}{2} a_s t^2
\end{align}
\label{eq:traj_const_acceleration}
\end{subequations}
where $V_0$ is an initial speed that, in the forward transition, is equal to zero. The reverse transition from cruising to hover can be described by setting $a_s$ negative and letting $V_0 = V_s$. 

The advantage of this trajectory is that it is computationally efficient, but unfortunately this method does not take into account the vehicle dynamics. This could be problematic for real implementation where instantaneous changes in acceleration are not feasible.  

\begin{figure}[!ht]
\makebox[0.99\linewidth][c]{
    \centering
    \begin{subfigure}{0.35\linewidth}
        \centering
        \includegraphics[width=1\linewidth,trim = {0.8cm 0 1cm 0}, clip]{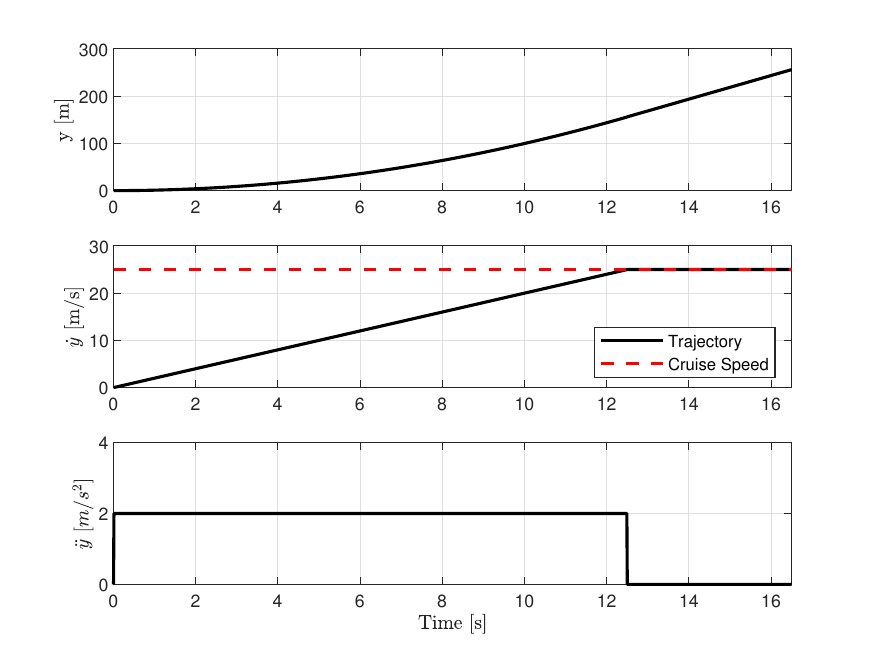}
        \caption{}
        \label{fig:const_acc_traj}
    \end{subfigure}
    \begin{subfigure}{0.35\linewidth}
        \centering
        \includegraphics[width=1\linewidth,trim = {1cm 0 1cm 0}, clip]{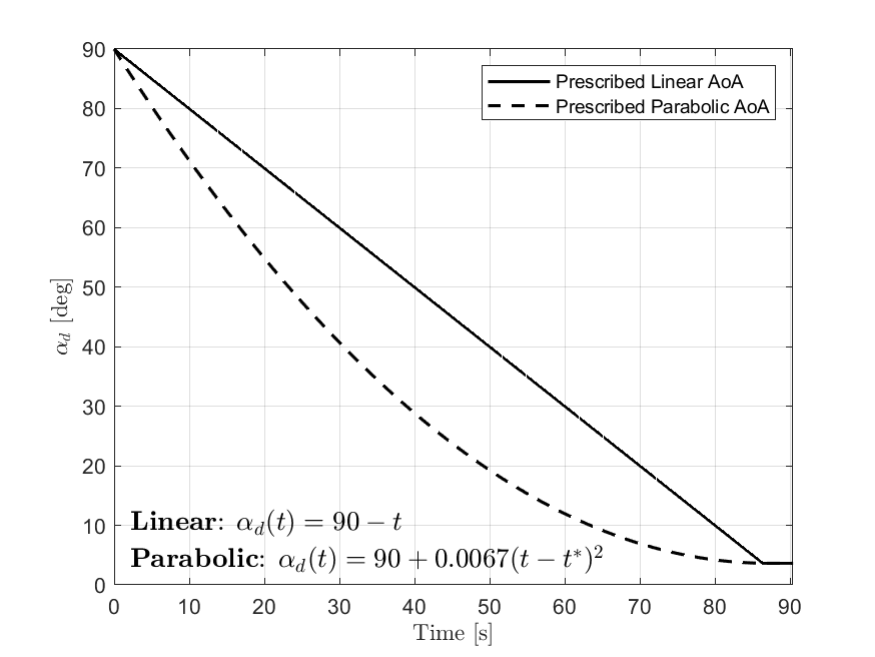}
        \caption{}
        \label{fig:prescribed_aoa_alpha_d}
    \end{subfigure}
    \begin{subfigure}{0.35\linewidth}
        \centering
        \includegraphics[width=1\linewidth,trim = {0.8cm 0 1cm 0}, clip]{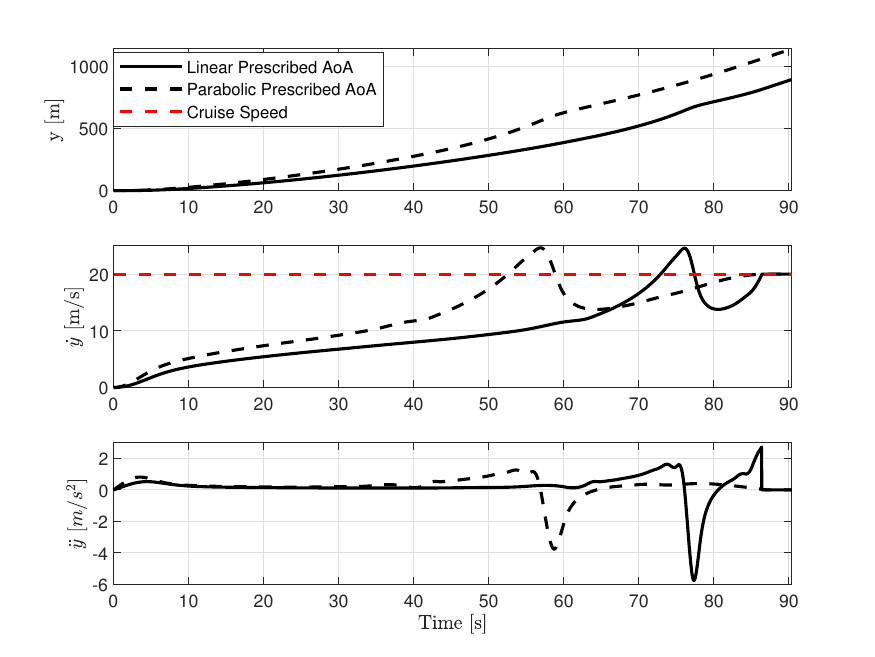}
        \caption{}
        \label{fig:prescribed_aoa_traj}
    \end{subfigure}
}
    \caption{\small \textbf{(a)} A trajectory generated using a desired constant acceleration; \textbf{(b)} Two examples of time-valued prescribed angle of attack functions for transition maneuvers with the parameters $\alpha_i = 90\degree, \alpha_f = 3.47\degree,$ and $t^* = 87$-sec; \textbf{(c)} Trajectories resulting from the mapping between airspeed and angle of attack (Equation \ref{eq:prescribed_aoa_diffeq}) and given the prescribed angle of attack functions from (b).}
    \label{fig:trajectories}
\end{figure}

\subsubsection{Prescribed angle of attack} \label{subsec:traj_prescribed_aoa}

Motivated by the shortcomings of the previous method, another trajectory was formulated that leverages the vehicle dynamics. This method was inspired by the existence of flight equilibria discussed in Section \ref{sec:stability_analysis}. 

In particular, the prescribed angle of attack method relies on a known mapping between the forward flight speed and equilibrium pitch angle; an example of this mapping is Figure \ref{fig:bifurcation_aoa}, where $a_v$ is directly proportional to the airspeed squared (this relationship is defined by Equation \ref{eq:a_v}). In that section, we observe that multiple equilibrium body orientations may exist for a desired airspeed. However, if we flip the axes, we can instead interpret Figure \ref{fig:bifurcation_aoa} as: "for a given equilibrium angle of attack, $\alpha_d$, there exists one and only one corresponding airspeed". In mathematics, the mapping $\alpha \mapsto a_v$ is considered \textit{surjective} or \textit{onto}. Critically, this is not true for the reverse mapping indicating we cannot get a unique angle of attack from a desired airspeed.

We first define a desired angle of attack as an arbitrary function of time, $\alpha_d = \alpha_d(t)$. There are no obvious restrictions on $\alpha_d(t)$ because the mapping is continuous and relatively smooth. Here we propose two functions for $\alpha_d(t)$ and assess their effectivenss in Section \ref{subsec:traj_prescribed_aoa}. The first function is a linear interpolation between two desired angles of attack: $\alpha_i , \alpha_f$, corresponding to the initial and final angles of attack, respectively. 
\begin{equation}
    \alpha_d(t) = \left\{
        \begin{array}{ll}
           \alpha_i - \Bigr ( \frac{\alpha_i - \alpha_f}{t^*} \Bigr ) t  & \quad t \leq t^* \\
            \alpha_f & \quad t > t^*
        \end{array}
    \right.
\label{eq:prescribed_aoa_linear}
\end{equation}
where $t^*$ is the target transition time in seconds. This time can be tuned for different objectives such as minimizing time in transition or satisfying actuation constraints. The second proposed function is a parabola also defined by $\alpha_i, \alpha_f,$ and the target transition time. 
\begin{equation}
    \alpha_d(t) = \left\{
        \begin{array}{ll}
           \alpha_f - \Bigr (\frac{\alpha_f - \alpha_i}{t^{*2}} \Bigr )(t-t^*)^2  & \quad t \leq t^* \\
            \alpha_f & \quad t > t^*
        \end{array}
    \right.
\label{eq:prescribed_aoa_parabola}
\end{equation}
Both the parabola and linear interpolation can be seen in Figure \ref{fig:prescribed_aoa_alpha_d} for $\alpha_i = 90\degree, \alpha_f = 3.47\degree,$ and $t^* = 87$-sec.

Given a prescribed angle of attack, the forward airspeed is extracted as a function of time based on the mapping between $\alpha$ and $a_v$. For a constant altitude transition maneuver, the airspeed is equal to $\dot{y}$. Initially, we may try to use the static relationship between $\alpha$ and $a_v$ expressed by Equation \ref{eq:a_v_aoa} to solve for the airspeed. But since this equation is derived at steady state, a trajectory generated from Equation \ref{eq:a_v_aoa} would not account for any body accelerations experienced during transition. 

Instead, we can leverage the aircraft's dynamics by starting with the forces projected onto the $\boldsymbol{\hat{b_3}}$ axis, as in Equation \ref{eq:stability_dynamics_b3}, and noting that for a constant altitude transition maneuver: $a_{b3} = -\ddot{y}(t) \sin \theta$, and in the absence of prop-wash, $\theta = \alpha$. In Section \ref{sec:stability_analysis}, $\boldsymbol{V_r}$ was a desired reference velocity with both a $y$ and $z$ component. For a constant altitude maneuver, $\boldsymbol{V_r}$ only has a $y$ component, which will be denoted by $\dot{y_r}(t)$. We can further simplify Equation \ref{eq:a_v_aoa} by assuming no prop-wash; i.e. $\alpha_e = \alpha$. The last step is to prescribe a desired angle of attack versus time, $\alpha = \alpha_d(t)$. The resulting expression is a first-order, nonlinear ordinary differential equation of $\dot{y_r}(t)$ with respect to time. 
\begin{equation}
    \ddot{y_r}(t) + \frac{\frac{1}{2}\rho S_{wing} \Bigr [ C_L(\alpha_d(t)) \cos \alpha_d(t) + C_D(\alpha_d(t)) \sin \alpha_d(t)   \Bigr]}{m \sin \alpha_d(t)} \dot{y_r}^2(t) = g \cot \alpha_d(t)
\label{eq:prescribed_aoa_diffeq}
\end{equation}
This nonlinear ODE is presented in the form: $\ddot{y_r}(t) + A(t) \dot{y_r}^2(t) = B(t)$ where the coefficients $A(t), B(t)$ are time-dependent precisely because of the prescribed angle of attack, $\alpha_d(t)$. The solution to Equation \ref{eq:prescribed_aoa_diffeq} above is a time-valued function, $\dot{y_r}(t)$, that satisfies both the dynamics of the QBiT and the desired time evolution of the angle of attack on the wing. With an expression for the horizontal airspeed, $\dot{y_r}(t)$, we can integrate and differentiate to get the position ($y_r(t)$) and acceleration ($\ddot{y_r}(t)$), respectively. The reverse transition is generated in a similar fashion, given the appropriate prescribed angle of attack. 

In summary, we have provided a method of generating dynamically feasible, constant altitude transition maneuvers that can, in theory, result in a desired time evolution of the angle of attack of the wings.  There are two unfortunate caveats to this method: 1) care must be taken in designing $\alpha_d(t)$ such that the solution to Equation \ref{eq:prescribed_aoa_diffeq} obeys physical limitations of the system (e.g. thrust limits), and 2) while the solution is built on the existence of equilibria, it does not account for the stability of those fixed points. As we will see in Section \ref{sec:prescribed_aoa_alt_transtion}, any small deviation from the unstable equilibria will push the QBiT to nearby stable ones. This method was formulated in 1-D, but future work will consider planar trajectories and the reduced-order prop-wash model, both of which could have added benefits of transitioning while enforcing constraints on angle of attack such as stall. 
\subsection{Simulation Environment} \label{sec:simulator}

Validation of the methods described above was performed with a simulation environment handwritten in MATLAB \cite{MATLAB:2019}. The point-mass dynamic model described in Section \ref{subsec:point-mass_dynamic_model} was simulated following trajectories taken from Section \ref{sec:traj_gen}, and the vehicle was stabilized by the controller formulated in Section \ref{sec:geometric_controller}. The overall structure and code of the simulation can be seen in the Appendix, Figure \ref{fig:sim_flowchart}.

Iterations occurred at a rate of 100 Hz ($dt = 0.01$ seconds) in order to minimize integration errors while also resembling the performance of typical microcontrollers available today. Numerical integration of the dynamics was performed using the 4th-order Runge-Kutta method which has a local truncation error on the order of $O(dt^5)$ \cite{RK4_method}. The physical parameters and controller gains used for results to follow are summarized in the Appendix, Table \ref{table:parameters}. 

\section{Results and Discussion}
In this section, we first use trim analysis of the dynamics to characterize the performance of a tailsitter across a variety of flight speeds, and identify the effects of the prop-wash model in steady-level flight. We then evaluate the controller's ability to track two constant-altitude method of transition from hover to forward flight. 

\subsection{Trim Analysis} \label{sec:trim_analysis}

Trim analysis is a useful tool for assessing the flight characteristics of an aircraft across its flight domain, such as how the aerodynamic forces or stable angle of attack varies with airspeed or flight path angle. Recall that trim flight is achieved when the aircraft is not accelerating. In the case of the QBiT, numerical trim analysis is particularly useful for studying the effects of the reduced-order prop-wash airflow model presented in Section \ref{subsec:ref_frames}, where an equilibrium solution for the angle of attack is quite difficult to solve for analytically.

In this section, we explore the effects that the prop-wash model has on the QBiT at steady-level flight. We numerically estimated the equilibrium angle of attack for airspeeds in the range $||\boldsymbol{V_i}|| \in [1, 30]$ incremented by $1$-m/s and prop-wash efficiencies $\eta \in [0,1]$ incremented by $0.05$ for a total of 630 trim points. The hand-built trim solver uses gradient descent to converge on the equilibrium thrust and body orientation, $\theta$, for each flight condition, and the tailsitter is simulated for 5 seconds using these initial conditions to let the system stabilize to steady-state flight. 

\begin{figure}[!ht]
    \centering
    \begin{subfigure}{0.48\linewidth}
        \centering
        \includegraphics[width=1\linewidth]{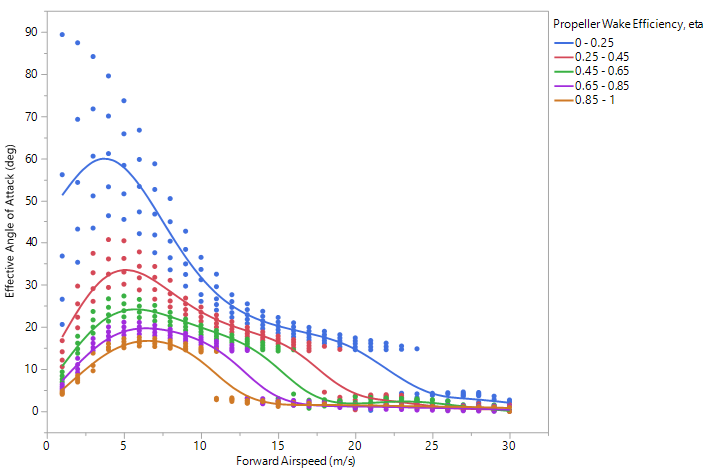}
        \caption{}
        \label{fig:trim_analysis_Vi}
    \end{subfigure}
    \begin{subfigure}{0.48\linewidth}
        \centering
        \includegraphics[width=1\linewidth]{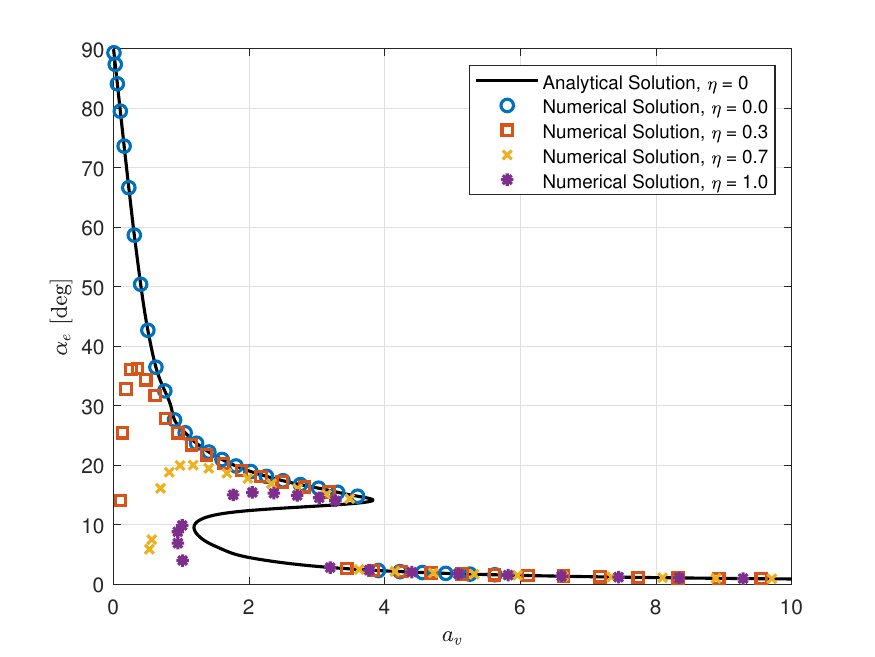}
        \caption{}
        \label{fig:trim_analysis_av}
    \end{subfigure}
    \caption{\small \textbf{(a)} The equilibrium angle of attack corresponding to a forward flight speed, $V_i$, plotted for varying prop-wash efficiency ranges; \textbf{(b)} The stable angle of attack varied against the aerodynamic loading, $a_v$, for select prop-wash efficiencies. These values are compared with the analytical expression for $a_v$ described in Equation \ref{eq:a_v_aoa}.}
    \label{fig:trim_analysis}
\end{figure}

The main results of the trim analysis are summarized in Figure \ref{fig:trim_analysis}. In Figure \ref{fig:trim_analysis_Vi}, the equilibrium angle of attack is plotted for steady-level flight, colored by a range of prop-wash efficiencies as indicated to the right of the plot. For each range of efficiencies, a smooth fit is applied to indicate the general trend of the equilibrium angles. In Figure \ref{fig:trim_analysis_av}, we plot the equilibrium angle of attack versus the aerodynamic loading, $a_v$, which is computed from Equation \ref{eq:a_v} using the true airflow speed, $||\boldsymbol{V_a}||$. These trends are compared with the analytical expression for $a_v$ as a function of the angle of attack derived in Equation \ref{eq:a_v_aoa} which does not include the prop-wash model. 

In Figure \ref{fig:trim_analysis_Vi}, we observe that the discontinuous jump in the equilibrium angle of attack occurs at increasing airspeed with decreasing prop-wash efficiency (orange to blue) indicative of an inverse relationship. The discontinuity itself is a consequence of the trim solver's gradient descent method, which never converges on the unstable region (such as that revealed in Section \ref{sec:stability_analysis}), opting for a nearby stable equilibrium instead. The location, or airspeed, at which discontinuity occurs is reflective of the relationship between the \textit{true airflow} over the wing as opposed to the inertial velocity of the aircraft. In this context, the aerodynamic loading is more appropriate to analyze discontinuity because it describes the airflow over the wing regardless of whether it is due to prop-wash or translation.

In Figure \ref{fig:trim_analysis_av}, we visualize the effect of prop-wash by comparing the numerically solved trim values to the analytical solution that neglects prop-wash. In contrast to Figure \ref{fig:trim_analysis_Vi}, the discontinuity occurs roughly at the same location ($a_v\approx3.82$) confirming that airflow, not inertial velocity, is the dominating factor in determining when this discontinuity occurs. Again we note that for $a_v<2$ the shape of the equilibria varies considerably between different prop-wash efficiencies. Beyond this value, however, all prop-wash conditions converge onto the analytical approximation as the inertial velocity dominates the airflow characteristics of the wing. The implication here is that at low speeds prop-wash can have significant effects on both the location and shape of equilibria. 

\subsection{Transition 1: Constant Acceleration} \label{sec:const_alt_transtion}
In the first transition maneuver, the vehicle attempts a forward transition using a trajectory derived from a constant acceleration. In this transition we assume no prop-wash and the flight path is horizontal, so $\alpha_e = \alpha = \theta$. The vehicle begins at hover with $\theta = 90\degree$ and accelerates forward at a rate of $2$-m/s$^2$ to a cruising airspeed of $25$-m/s. The simulation continues for an additional 4 seconds to capture the response from the controller. 

\begin{figure}[!ht]
    \centering
    \begin{subfigure}{0.48\linewidth}
        \centering
        \includegraphics[width=1\linewidth]{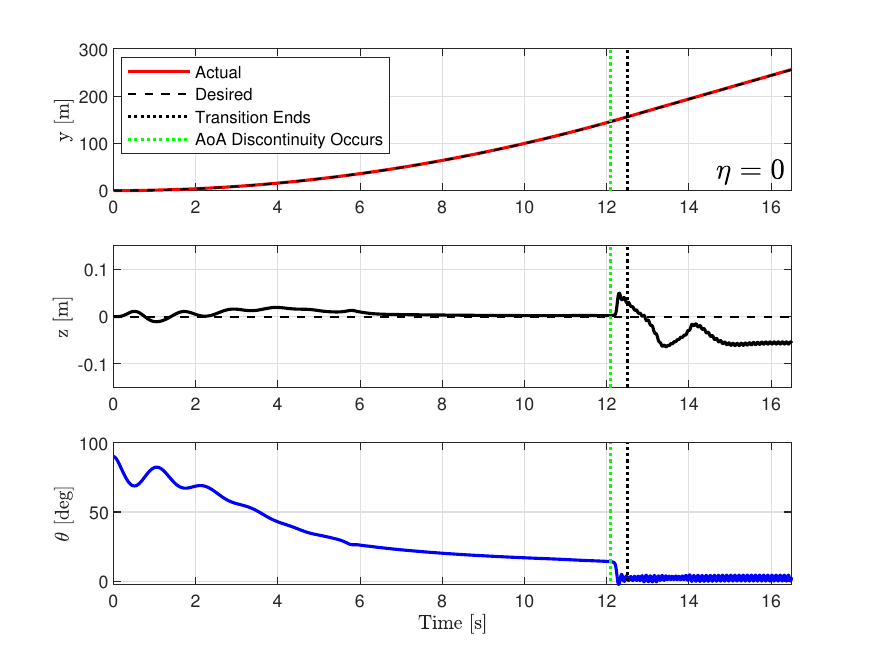}
        \caption{}
        \label{fig:results_const_acc_states}
    \end{subfigure}
    \begin{subfigure}{0.48\linewidth}
        \centering
        \includegraphics[width=1\linewidth]{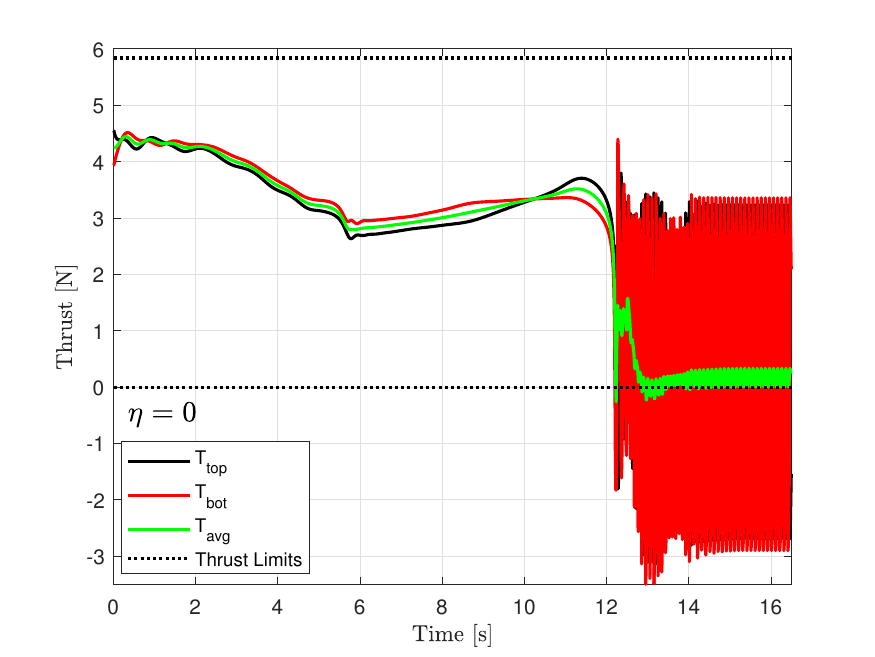}
        \caption{}
        \label{fig:results_const_acc_thrusts}
    \end{subfigure}
    \caption{\small Results for a constant altitude transition maneuver occurring with a constant forward acceleration of 2-m/s. \textbf{(a)} The states indicating the position and body orientation of the QBiT throughout the transition. The green dotted line indicates when a jump in the equilibrium angle of attack occurs, and the black dotted line marks the start of forward cruise; \textbf{(b)} The thrusts on the top and bottom wings produced by the controller in response to the trajectory. The dotted lines indicate approximate thrust limits for a set of propellers.}
    \label{fig:results_const_acc}
\end{figure}

The states and thrust inputs are plotted in Figure \ref{fig:results_const_acc}. The transition maneuver takes only $12.5$-sec and requires approximately 150-m of horizontal space. The maximum error in y and z from the trajectory are $0.24$-m and $0.06$-m, respectively, indicating reasonable tracking abilities for this maneuver. However, at $t = 12.1$-sec, the vehicle experiences a sudden discontinuous jump in the equilibrium angle of attack from $\theta = 14.1\degree$ to $\theta \approx 2.33\degree$. This discontinuity sparks a very large response from the controller and the vehicle never truly recovers. 

The jump in pitch angle can be explained by the bifurcation phenomena existing in the equilibria angle of attack. In transition, the discontinuity occurs when $a_v \approx 3.82$ ($||\boldsymbol{V_i}|| = 24$-m/s) which is notably where the number of equilibria solutions reduces from 3 to 1 as seen in Figure \ref{fig:bifurcation_aoa}. Since the vehicle is always accelerating, $a_v$ can never decrease, and therefore the equilibrium pitch angle jumps down to the single remaining equilibrium: $2.33\degree$. In this particular scenario, the controller response is large enough to exceed the lower bound on the thrust for both wings. In reality this maneuver would likely cause a loss of control and presents a serious danger for VTOL aircraft attempting level transition.

\subsection{Transition 2: Prescribed Angle of Attack} \label{sec:prescribed_aoa_alt_transtion}
In the constant acceleration trajectory, we saw a large discontinuity in the pitch angle that would likely lead to a grounded aircraft. The method described in Section \ref{sec:prescribed_aoa_alt_transtion} could perhaps mitigate this by enforcing a continuous angle of attack. Below we employ the parabola trajectory (which can be seen in Figure \ref{fig:prescribed_aoa_alpha_d}) in an attempt to enforce a continuous pitch angle throughout a transition to the cruising speed of $25$-m/s. As was the case with the previous attempt, prop-wash is ignored and the transition occurs at a constant altitude.

\begin{figure}[!ht]
    \centering
    \begin{subfigure}{0.48\linewidth}
        \centering
        \includegraphics[width=1\linewidth]{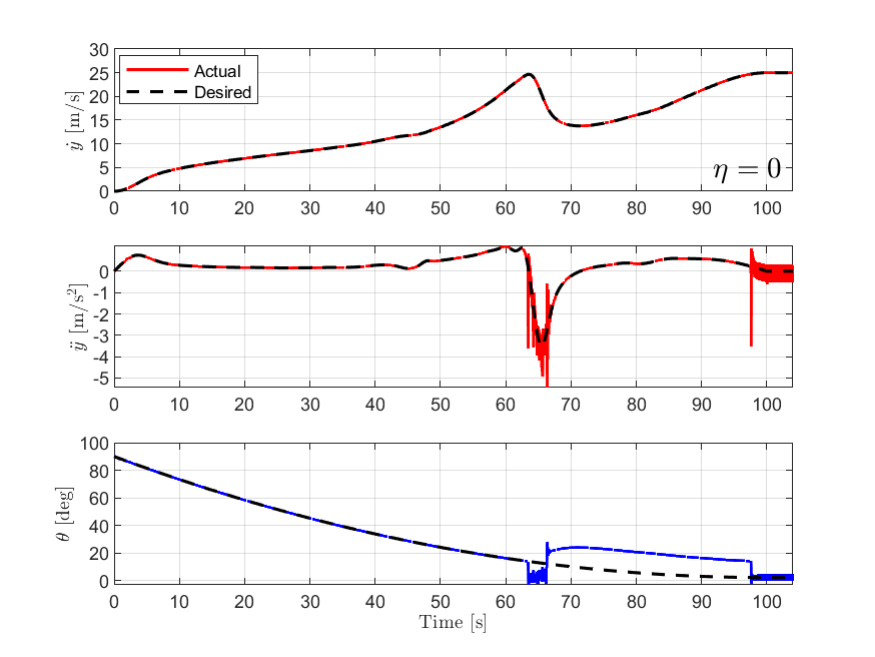}
        \caption{}
        \label{fig:results_prescribed_aoa_states}
    \end{subfigure}
    \begin{subfigure}{0.48\linewidth}
        \centering
        \includegraphics[width=1\linewidth]{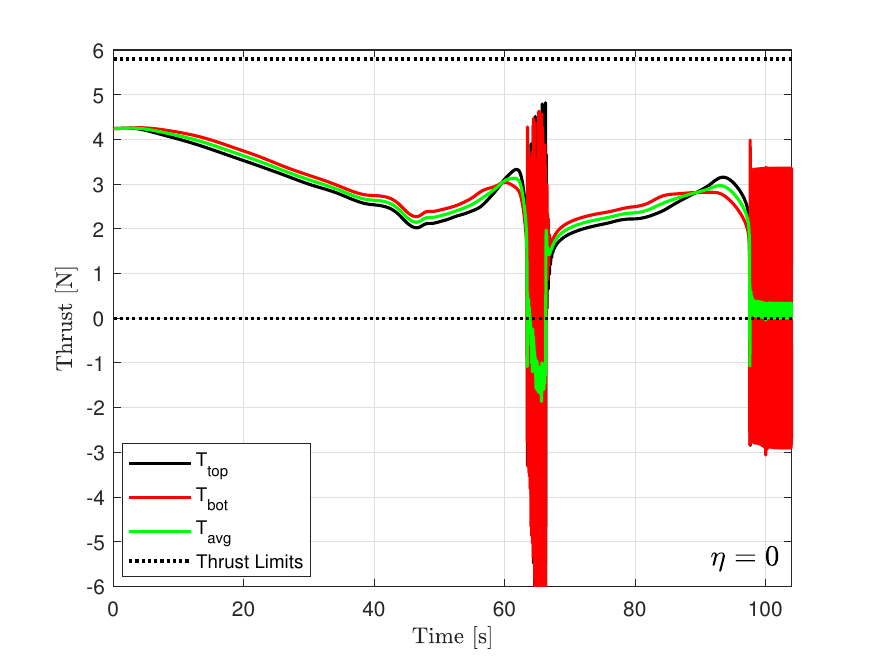}
        \caption{}
        \label{fig:results_prescribed_aoa_thrusts}
    \end{subfigure}
    \caption{\small Results for a transition maneuver using a parabolic desired angle of attack function \textbf{(a)} The horizontal speed, acceleration, and pitch angle compared to the trajectory values; \textbf{(b)} The thrusts on the top and bottom wings produced by the controller in response to the trajectory. The dotted lines represent thrust limits for the motors on each wing.}
    \label{fig:results_prescribed_aoa}
\end{figure}

The results for this trajectory were below satisfactory. The transition takes over 100 seconds to complete and requires over $1.3$-km of horizontal airspace. This is a consequence of trying to minimize the harsh negative acceleration necessary for slowing the vehicle down to match the changing equilibria. The maximum y and z position errors for this maneuver are $0.25$-m and $0.15$-m, respectively. The altitude error is 3x larger than that in the constant acceleration case. The matter worsens when assessing the controller's response: between time $t \in [65,69]$, corresponding to a sudden negative acceleration, the thrust is sporadic and unpredictable as the controller tries to stabilize the vehicle. While pitch tracking is very good initially ($error[\theta] < 0.11\degree)$ for $t<65$), the error jumps to roughly 12$\degree$ for $t\in[65,97]$. At $t=97$ the pitch angle seems to settle back to the desired angle of attack but requires a large step response to do so. 

Despite the shortcomings, this method has some useful insights. As designed, the negative acceleration decreases $a_v$ in order track the equilibria, stable or not, that coincides with the desired angle of attack. As we noted in Section \ref{sec:stability_analysis}, the equilibria in the region $\alpha \in [10,14]$ are unstable. As such, what we are actually observing is the controller endlessly failing to stabilize to an unstable equilibria, instead settling to stable equilibria until time $t = 97$ when there are no longer multiple solutions. This phenomenon highlights the need for methods to stabilize any desired pitch angle either through the addition of control surfaces or drawing inspiration from similar problems in classical nonlinear control. 

This method does have two advantages over the constant acceleration trajectory. We no longer see step responses at the beginning of the transition maneuver because the acceleration is continuous. This could perhaps avoid conditions leading to loss of control at the beginning of transition. The second advantage is that we have much more control of the time evolution of the wing's angle of attack with this method, so long as we do not cross the unstable region of equilibria. If we were to remove the constant altitude constraint and apply this method, it is possible to design a 2-D transition maneuver that obeys stall constraints on the angle of attack without the need for numerical optimization. 

\section{Conclusions and Future Work}
In this paper, we considered problems related to the transition maneuvers of hybrid VTOL for applications in package delivery. We first derived equations of motion for a reduced-order dynamic model of a generic thrust-actuated tailsitter. We performed a stability analysis of the vehicle for tracking reference velocities, then outlined a nonlinear geometric controller that stabilizes the QBiT to an arbitrary trajectory. Two methods of trajectory generation were described, the latter being a novel attempt to leverage the unique mapping between a winged VTOL's angle of attack and its airspeed in 1-D. These trajectories were then evaluated in a hand-built simulation environment to assess the potential for real-world implementation.

Results from this study indicate that even for low accelerations, forward transition is very difficult: the existence and nature of equilibrium angles of attack can result in a discontinuous jump for a constant altitude maneuver if the controller does not actively handle the stability of an equilibrium. The bifurcation of equilibria presents a real danger for aircraft attempting to operate in the post-stall regime, and this work clearly demonstrates its effect on the transition maneuver in a controlled simulation. An attempt to force a continuous time evolution for the pitch angle, essentially avoiding a discontinuity due to bifurcation, failed for two reasons: the vehicle lacks control surfaces that can stabilize arbitrary equilibria, and the controller relies on the passive stability of an equilibria to stabilize the aircraft. 

This work has produced indicators that constant altitude transition with a continuously-defined pitch angle is possible, albeit difficult and nuanced. In future work, we will consider what types of functions for the prescribed angle of attack might lend itself to stable and continuous transition. More importantly, we will consider ways to stabilize the unstable equilibria using classical approaches from nonlinear control (e.g. energy shaping) that may deviate from the geometric controller presented in this work. On the planning side, we will work to formulate this trajectory generation method in 2-D. At the cost of unconstrained altitude behavior, a 2-D formulation of this method could provide a transition satisfying constraints of angle of attack in a computationally efficient fashion. Lastly, we will make efforts towards integrating the prop-wash model into this framework, further expanding the stability of the aircraft by leveraging its effect on the angle of attack. The implications of this work, present and future, are more efficient, predictable, and capable UAVs for use in autonomous package delivery in constrained environments. 

\section*{Acknowledgements}
\addtocounter{section}{1}

The author would like to acknowledge the advising of Dr. Mark Yim, Dr. Vijay Kumar, and Dr. M. Ani Hsieh on this project and throughout the past year. He would also like to thank all his peers who provided input and helped prepare him for this examination, in particular Greg Campbell, Parker Lamascus, and Jake Welde. The author would also like to acknowledge Dr. Jean-Paul Reddinger, Dr. John Gerdes, and Dr. Michael Avera from the U.S. Army Research Laboratory for early input and providing physical parameters for simulation. 

\bibliography{mybib}{}
\bibliographystyle{unsrt}

\newpage
\appendix
\appendixpage
\addappheadtotoc
\section{Controller Step Responses}
\subsection*{Hover}
\begin{figure}[!ht]
    \centering
    \includegraphics[width=1\linewidth]{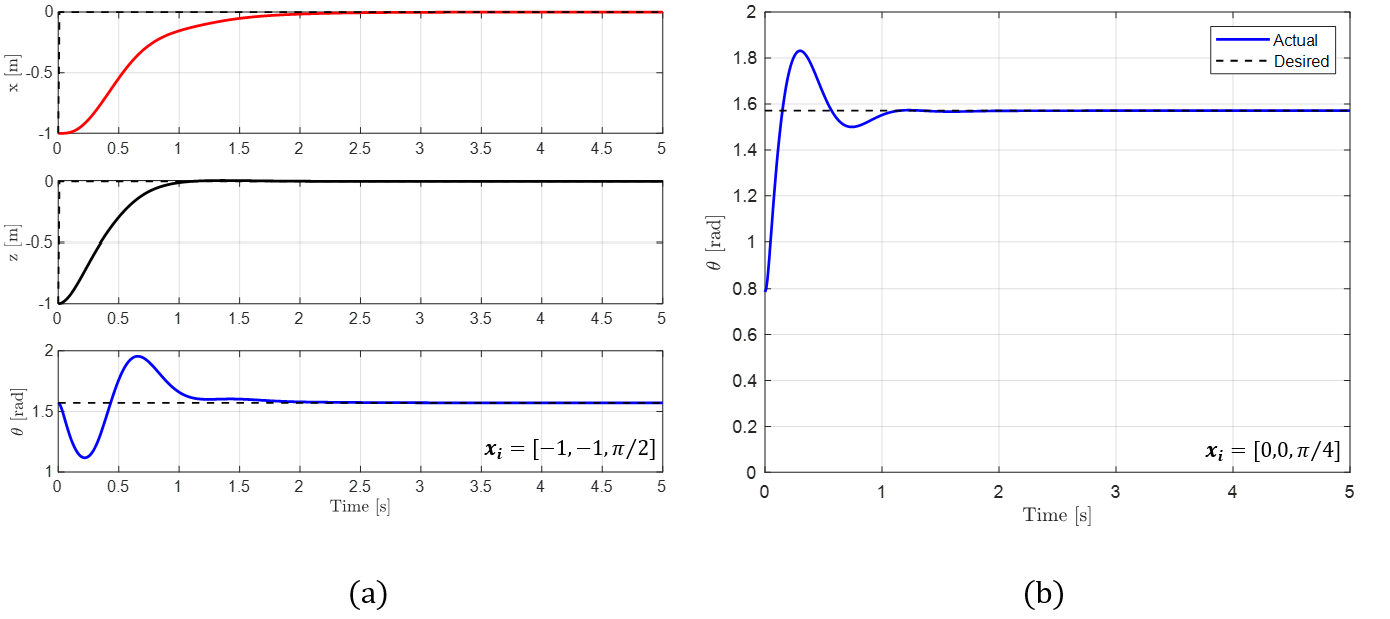}
    \caption{\small Step responses of the geometric controller at hover for \textbf{(a)} a position error of $\Delta y = -1$ and $\Delta z = -1$ with an approximate settling time of 1.5-sec; \textbf{(b)} an attitude error of $\Delta \theta = -pi/4$ and an approximate settling time of 1-sec.}
    \label{fig:step_response_hover}
\end{figure}

\subsection*{Cruising}
\begin{figure}[!ht]
    \centering
    \includegraphics[width=1\linewidth]{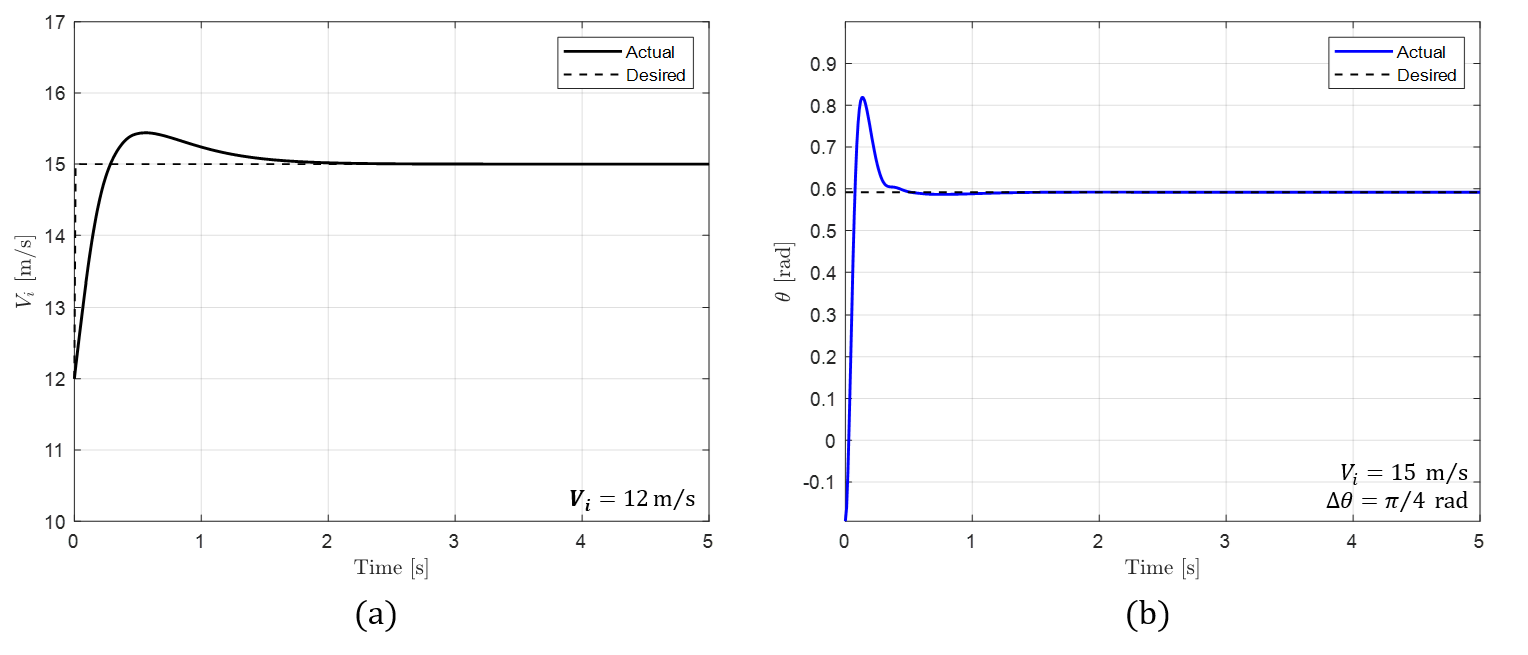}
    \caption{\small Step responses of the geometric controller at cruise for \textbf{(a)} an initial speed error of $\Delta V_i = -3 m/s$ with a corresponding settling time of approximately 1.5-sec; \textbf{(b)} an initial attitude error of $\Delta \theta = \pi/4$ with a settling time of roughly 0.2-sec.}
    \label{fig:step_response_cruise}
\end{figure}

\section{Data Sets}
\subsection*{Transition - constant acceleration}
\begin{figure}[!h]
    \centering
    \includegraphics[width=0.9\linewidth]{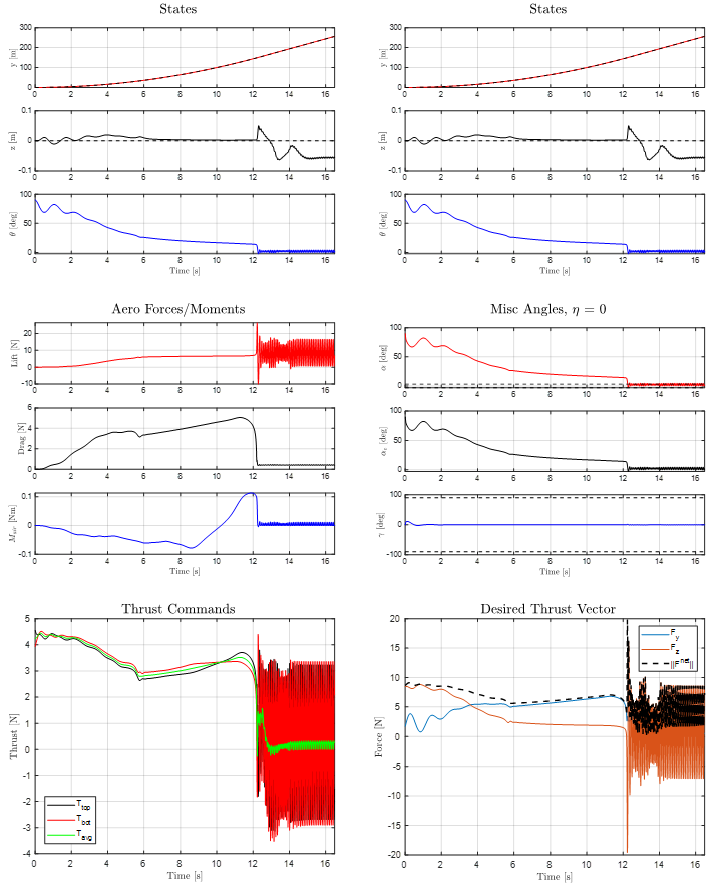}
    \caption{\small The full sets of figures for a constant-acceleration forward transition. This includes the states, state derivatives, aerodynamic forces and pitching moment, angles of attack, flight path angle, and controller outputs.}
    \label{fig:results_const_acc_all}
\end{figure}

\subsection*{Transition - prescribed AoA}
\begin{figure}[!h]
    \centering
    \includegraphics[width=0.9\linewidth]{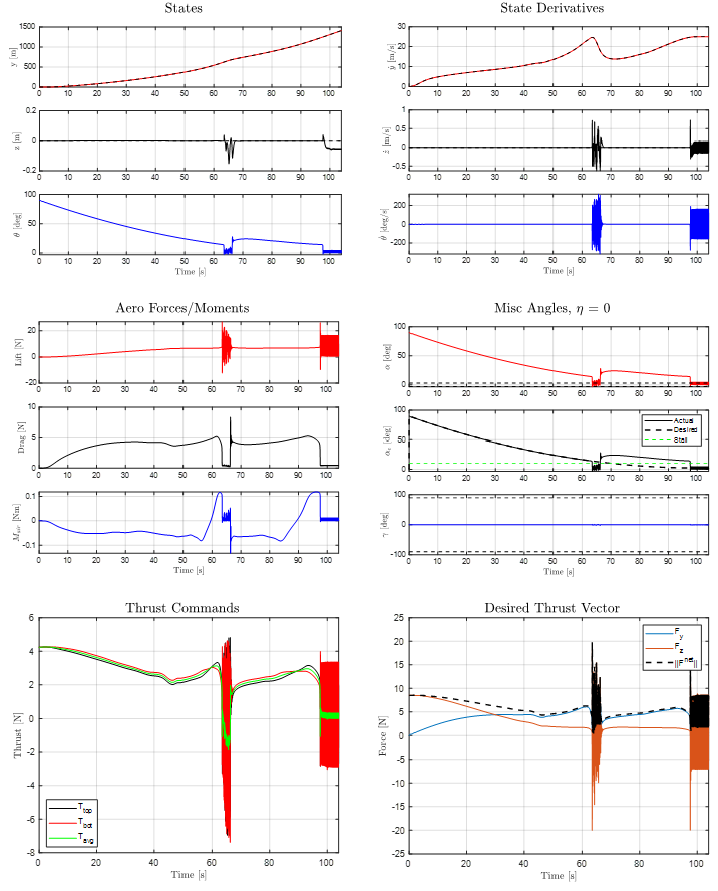}
    \caption{\small The full sets of figures for a forward transition with a prescribed parabolic angle of attack. This includes the states, state derivatives, aerodynamic forces and pitching moment, angles of attack, flight path angle, and controller outputs.}
    \label{fig:results_prescribed_aoa_full}
\end{figure}

\section{Simulation Environment}

\subsection*{Code Structure and Parameters}
\begin{figure}[!ht]
    \centering
    \includegraphics[width=0.85\linewidth]{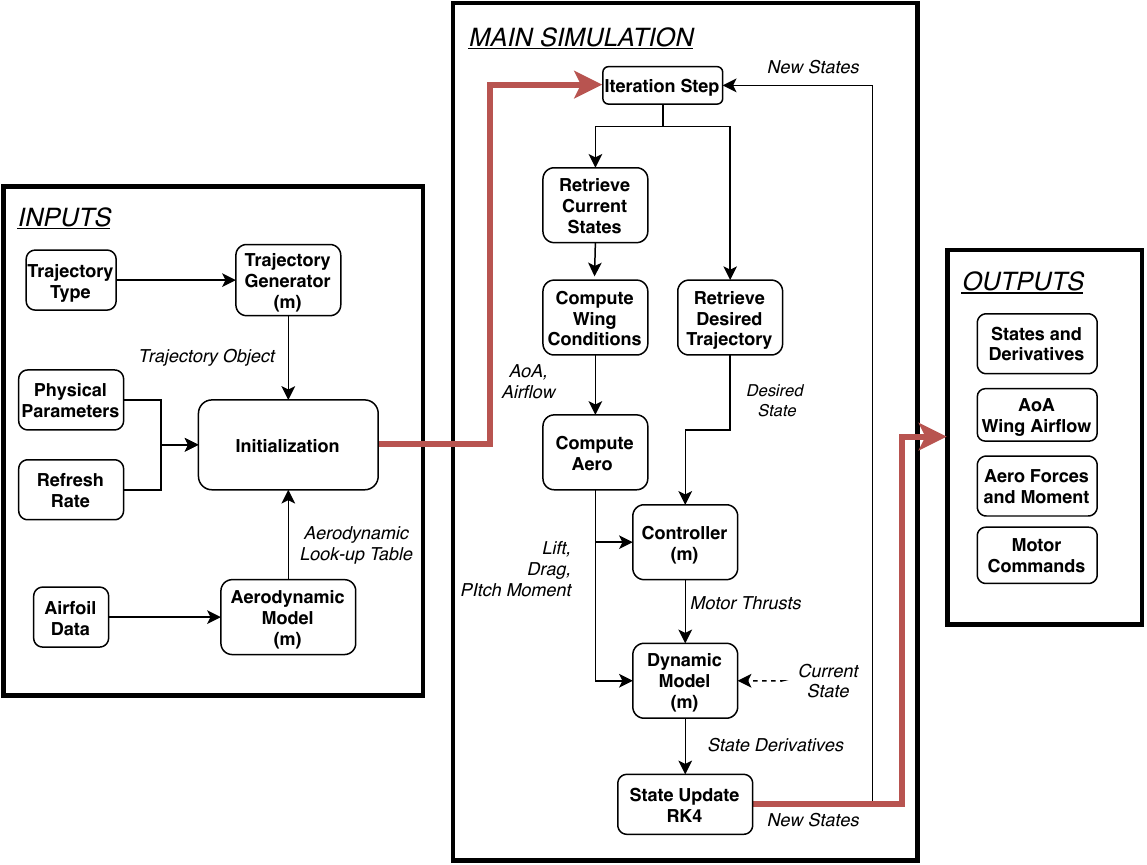}
    \caption{A flowchart indicating the sequence of events and flow of information for the main simulation script in MATLAB. Blocks with a (m) sign indicate separate MATLAB files}
    \label{fig:sim_flowchart}
\end{figure}

\begin{table}[!ht]
\centering
    \begin{tabular}{|c|c|c|}
    \hline
        \textbf{Parameter} & \textbf{Value} & \textbf{Units} \\
    \hline \hline
    \multicolumn{3}{|c|}{\textit{Physical Parameters}} \\
    \hline
    Mass ($m$) & 0.8652 & kg \\
    Inertia ($I_{yy}) $ & 9.77E-03 & kg-m$^2$ \\
    Arm Length ($l$) & 0.244 & m \\
    Wing Chord ($\bar{c}$) & 0.087 & m \\
    Wing Span ($b$) & 1.016 & m \\
    Rotor Diameter ($2R$) & 0.229 & m \\
    Min Thrust ($T_{min}$) & 0 & N \\
    Max Thrust - 2 Motors ($T_{max}$) & 5.886 & N \\
    Thrust-Weight Ratio ($TW$) & 1.387 & - \\
    \hline
    \multicolumn{3}{|c|}{\textit{Controller Gains}} \\
    \hline
    Proportional - Position ($K_p$) & \textit{diag}(11.6,17.4) & m$^{-1}$ \\
    Derivative - Position ($K_d$) & \textit{diag}(6.82,6.82) & s-m$^{-1}$ \\
    Proportional - Attitude ($K_R$) & 74.73 & rad$^{-1}$ \\
    Derivative - Attitude ($K_{\omega}$) & 17.29 & s-rad$^{-1}$ \\
    \hline
    \end{tabular}
\caption{\small A summary of the physical parameters and controller gains used in the simulation of transition maneuvers}
\label{table:parameters}
\end{table}

\subsection*{Code}
\begin{lstlisting}[language=Octave,caption={The main MATLAB file the handles initialization, trajectory generation, dynamics, and plotting.}, captionpos=t]
%%% Simulating the dynamics of the qbit. This script will establish state
%%% variables, get a trajectory, input that trajectory into a controller to
%%% get commands, and simulate the dynamics subject to those inputs.
%%% Spencer Folk 2020

clear
clc
close all

% Bools / Settings
aero = true;  % This bool determines whether or not we compute aerodynamic forces
animate = false; % Bool for making an animation of the vehicle.
save_animation = false; % Bool for saving the animation as a gif
integrate_method = "rk4";  % Type of integration - either 'euler' or 'rk4'
traj_type = "prescribed_aoa"; % Type of trajectory:
%                           "cubic",
%                           "trim" (for steady state flight),
%                           "increasing" (const acceleration)
%                           "decreasing" (const decelleration)
%                           "prescribed_aoa" (constant height, continuous AoA)
%                           "stepP" (step response in position at hover)
%                           "stepA_hover" (step response in angle at hover)
%                           "stepV" (step response in airspeed at trim)
%                           "stepA_FF" (step response in angle at forward flight)

%% Initialize Constants
in2m = 0.0254;
g = 9.81;
rho = 1.2;
stall_angle = 10;  % deg, identified from plot of cl vs alpha
dt = 0.01;   % Simulation time step

eta = 0.0;   % Efficiency of the down wash on the wings from the propellers

linear_acc = 2;   % m/s^2, the acceleration/decelleration used in
%                  "increasing" and "decreasing" trajectories
angular_vel = -1;   % deg/s, the desired change in attitude used by the
%                  "prescribed_aoa" trajectory
V_s = 25;          % m/s, set velocity used in "increasing", "decreasing", and
%                  "trim" trajectories...
end_time = 5;     % Duration of trajectory, this will be REWRITTEN by all but
%                  "trim" and "step___" trajectories.

step_angle = -pi/4; % the angular step used by stepA_hover (positive counter clockwise)
step_y = -1;          % step in the x direction used by stepP
step_z = -1;          % step in the z direction used by stepP
step_V = -3;          % step in forward airspeed used by stepV

buffer_time = 4;  % s, sim time AFTER transition maneuver theoretically ends
%                   ... this is to capture settling of the controller

%% Vehicle Parameters
% Load in physical parameters for the qbit

% CRC 5in prop
% m_airframe = 0.215;
% m_battery = 0.150;
% m = m_airframe + m_battery;
%
% Ixx = 2.32e-3;
% span = 15*in2m;
% l = 6*in2m;
% chord = 5*in2m;
% R = 2.5*in2m;

% CRC 9in prop (CRC-3 from CAD)
% Compute a scaling factor based on change in wing span:
span = 2*0.508;  % Doubled for biplane set up
l = 0.244;
chord = 0.087;
R = 4.5*in2m;   % Estimated 9in prop

scaling_factor = span/(15*in2m);
% m = (0.3650)*(scaling_factor^3);  % Mass scales with R^3
m = 0.8652;  % This is the value of expression above^ but we want it fixed
%              so we can change the span without worry
% Ixx = (2.32e-3)*(scaling_factor^5);
Ixx = 0.009776460905350; % This is the value of expression above^ but we want it fixed
%                          so we can change the span without worry

%% Generate Airfoil Look-up
% This look up table data will be used to estimate lift, drag, moment given
% the angle of attack and interpolation from this data.
[cl_spline, cd_spline, cm_spline] = aero_fns("naca_0015_experimental_Re-160000.csv");

%% Trajectory Generation
% Generate a trajectory based on the method selected. If cubic, use cubic
% splines. If trim, create a constant speed, trim flight.

if traj_type == "cubic"
    %     waypoints = [0,40; 0,0];
    % waypoints = [0,0,10 ; 0,10,10];  % aggressive maneuver
    % waypoints = [0,20,40 ; 0,0,0];  % Straight line horizontal trajectory
    waypoints = [0,80,160 ; 0,0,0];  % Straight line horizontal trajectory, longer
    % waypoints = [0,0,0 ; 0, 20, 40]; % Straight line vertical trajectory
    % waypoints = [0,10,40 ; 0,10,10];  % Larger distance shows off lift benefit
    % waypoints = [0,20,40 ; 0,5,10]; % diagonal
    % waypoints = [0,10,20,30,40 ; 0,10,0,-10,0]; % zigzag
    % waypoints = [0,0 ; 0, -10];  % Drop
    % waypoints = [0,0 ; 0, 10];  % rise
    
    [traj_obj, end_time] = qbit_spline_generator(waypoints, V_s);
    
    % Use this traj_obj to get our desired y,z at a given time t
    traj_obj_dot = fnder(traj_obj,1);
    traj_obj_dotdot = fnder(traj_obj,2);
    
    init_conds = [m*g/2; m*g/2 ; pi/2];
    
    % Time vector
    t_f = end_time;
    time = 0:dt:t_f;
    
    fprintf("\nTrajectory type: Cubic Spline")
    fprintf("\n-----------------------------\n")
    
elseif traj_type == "trim"
    % In the trim mode, we have to have a good initial guess for the trim
    % condition, so that the QBiT isn't too far from the steady state value
    % at the beginning of the trajectory!
    
    % This involves solving for T_top(0), T_bot(0), theta(0)
    x0 = [m*g/2; m*g/2; pi/4];
    fun = @(x) trim_flight(x, cl_spline, cd_spline, cm_spline, m,g,l, chord, span, rho, eta, R, V_s);
    %     options = optimoptions('fsolve','Display','iter');
    options = optimoptions('fsolve','Display','none','PlotFcn',@optimplotfirstorderopt);
    [init_conds,~,~,output] = fsolve(fun,x0,options);
    
    %     output.iterations
    
    % Time vector
    t_f = end_time;
    time = 0:dt:t_f;
    
    fprintf("\nTrajectory type: Trim")
    fprintf("\n---------------------\n")
    fprintf("\nTrim estimate solved: \n")
    fprintf("\nT_top = %3.4f",init_conds(1))
    fprintf("\nT_bot = %3.4f",init_conds(2))
    fprintf("\ntheta   = %3.4f\n",init_conds(3))
    
    waypoints = [0 , V_s*end_time ; 0, 0];
elseif traj_type == "increasing"
    % In this mode we use a constant acceleration to go from hover to V_s.
    % Therefore just set the initial condition to 0.
    
    init_conds = [m*g/2; m*g/2 ; pi/2];
    V_end = V_s;
    a_s = linear_acc;  % m/s^2, acceleration used for transition
    
    end_time = V_end/a_s + buffer_time;
    
    % Time vector
    t_f = end_time;
    time = 0:dt:t_f;
    
    fprintf("\nTrajectory type: Linear Increasing")
    fprintf("\n----------------------------------\n")
    
elseif traj_type == "decreasing"
    % Constant deceleration from some beginning speed, V_start, to hover.
    
    % Need to solve for an estimate of trim flight:
    x0 = [m*g/2; m*g/2; pi/4];
    fun = @(x) trim_flight(x, cl_spline, cd_spline, cm_spline, m,g,l, chord, span, rho, eta, R, V_s);
    %     options = optimoptions('fsolve','Display','iter');
    options = optimoptions('fsolve','Display','none','PlotFcn',@optimplotfirstorderopt);
    [init_conds,~,~,output] = fsolve(fun,x0,options);
    
    V_start = V_s;
    a_s = linear_acc;   % m/s^2, decelleration used for transition
    end_time = V_start/a_s + buffer_time;
    
    % Time vector
    t_f = end_time;
    time = 0:dt:t_f;
    
    fprintf("\nTrajectory type: Linear Decreasing")
    fprintf("\n----------------------------------\n")
    
elseif traj_type == "prescribed_aoa"
    % If it's constant height, design a desired AoA function
    % return a corresponding v(t), a(t), y/z(t) from that.
    
    % Need to solve for an estimate of trim flight:
    %     x0 = [m*g/2; m*g/2; 0];
    %     fun = @(x) trim_flight(x, cl_spline, cd_spline, cm_spline, m,g,l, chord, span, rho, eta, R, V_s);
    %     %     options = optimoptions('fsolve','Display','iter');
    %     options = optimoptions('fsolve','Display','none');
    %     [init_conds,~,~,output] = fsolve(fun,x0,options);
    
    a_v = 1/2*rho*chord*span*V_s^2/(m*g);
    x0 = 1e-3;
    options = optimoptions('fsolve','Display','none');
    
    fun = @(x) a_v - cot(x)/(ppval(cd_spline, x*180/pi) + ppval(cl_spline, x*180/pi)*cot(x));
    
    [init_conds,~,~,output] = fsolve(fun, x0, options);
    
    %%%%%%%%%%%%%%%%%%%%%%%%%%%%%%%%%%%%%%%%%%%%%%%%%%%
    % Constructing alpha_des:
    alpha_f = init_conds(end);  % Final value for alpha_des
    alpha_i = pi/2;  % Initial value for alpha_des
    
    alpha_traj_type = "parabolic";
    if alpha_traj_type == "linear"
        aoa_rate = angular_vel*(pi/180);  % Rate of change of AoA, first number in degrees
        end_time = abs(alpha_f - alpha_i)/abs(aoa_rate) + buffer_time;
        t_f = end_time;
        time = 0:dt:t_f;
        
        alpha_des = alpha_i + aoa_rate*time;
        alpha_des(time>(end_time-buffer_time)) = alpha_f;
    elseif alpha_traj_type == "parabolic"
        vertex_time = 87;    % seconds, time location of the vertex of parabola
        end_time = vertex_time + buffer_time;  % seconds
        t_f = end_time;
        time = 0:dt:t_f;
        
        a_coeff = (alpha_i - alpha_f)/((end_time-buffer_time)^2);
        alpha_des = alpha_f + a_coeff*(time-(end_time-buffer_time)).^2;
        alpha_des(time>(end_time-buffer_time)) = alpha_f;
        
    elseif alpha_traj_type == "decay"
        aoa_tau = 5;                        % seconds, time constant of first order decay
        %                                     used in alpha_traj_type = "decay"
        end_time = 4*aoa_tau + buffer_time; % seconds, we choose this.
        t_f = end_time;
        time = 0:dt:t_f;
        
        alpha_des = alpha_f + (alpha_i - alpha_f)*exp(-time./aoa_tau);
    end
    
    % Get temp trajectory variables and save them
    accel_bool = true;  % Consider acceleration when generating the trajectory
    [y_des, ydot_des, ydotdot_des]=prescribed_aoa_traj_generator(dt,time,alpha_des,cl_spline, cd_spline,rho,m,g,chord,span, accel_bool);
    
    fprintf("\nTrajectory type: Prescribed AoA")
    fprintf("\n-------------------------------\n")
    
elseif traj_type == "stepP" || traj_type == "stepA_hover"
    % For step hover, this is easy, we just need to set our trajectory to
    % zeros for all time
    time = 0:dt:end_time;
    
    fprintf("\nTrajectory type: Step Response at Hover")
    fprintf("\n---------------------------------------\n")
    
elseif traj_type == "stepV" || traj_type == "stepA_FF"
    % For the step in airspeed, we need to first set trim just like "trim"
    x0 = [m*g/2; m*g/2; pi/4];
    fun = @(x) trim_flight(x, cl_spline, cd_spline, cm_spline, m,g,l, chord, span, rho, eta, R, V_s);
    %     options = optimoptions('fsolve','Display','iter');
    options = optimoptions('fsolve','Display','none','PlotFcn',@optimplotfirstorderopt);
    [init_conds,~,~,output] = fsolve(fun,x0,options);
    
    %     output.iterations
    
    % Time vector
    t_f = end_time;
    time = 0:dt:t_f;
    
    fprintf("\nTrajectory type: Step in Flight")
    fprintf("\n-------------------------------\n")
else
    error("Incorrect trajectory type -- check traj_type variable")
end

fprintf(strcat("Integration Method: ",integrate_method));
fprintf(strcat("\nStep size: ",num2str(dt),"-sec"));
fprintf("\n----------------------------------\n")

%% Initialize Arrays

%%% TIME IS INTITALIZED IN THE SECTION ABOVE

% States
y = zeros(size(time));
z = zeros(size(time));
theta = zeros(size(time));

ydot = zeros(size(time));
zdot = zeros(size(time));
thetadot = zeros(size(time));

ydotdot = zeros(size(time));
zdotdot = zeros(size(time));
thetadotdot = zeros(size(time));

% Inputs

if traj_type == "trim" || traj_type == "decreasing"
    T_top = init_conds(1)*ones(size(time));
    T_bot = init_conds(2)*ones(size(time));
    
else
    T_top = m*g*ones(size(time));
    T_bot = m*g*ones(size(time));
end
% Misc Variables (also important)
alpha = zeros(size(time));
alpha_e = zeros(size(time));
gamma = zeros(size(time));

L = zeros(size(time));
D = zeros(size(time));
M_air = zeros(size(time));

Vi = zeros(size(time));
Va = zeros(size(time));
Vw = zeros(size(time));

% Bookkeeping the airflow over the top and bottom wings
Vw_top = zeros(size(time));
Vw_bot = zeros(size(time));

Fdes = zeros(2,length(time));  % Desired force vector

% Power consumption
Ptop = zeros(size(time));
Pbot = zeros(size(time));

% Initial conditions:
theta(1) = pi/2;
y(1) = 0;
z(1) = 0;
if traj_type == "trim" || traj_type == "decreasing"
    ydot(1) = V_s;
    theta(1) = init_conds(3);
elseif traj_type == "stepV"
    ydot(1) = V_s + step_V;
    theta(1) = init_conds(3);
elseif traj_type == "stepA_FF"
    ydot(1) = V_s;
    theta(1) = init_conds(3) + step_angle;
elseif traj_type == "stepP"
    y(1) = step_y;
    z(1) = step_z;
elseif traj_type == "stepA_hover"
    theta(1) = pi/2 + step_angle;
end
zdot(1) = 0;

% Trajectory state
desired_state = zeros(6,length(time));  % [y, z, ydot, zdot, ydotdot, zdotdot]
desired_state(:,1) = [y(1);z(1);ydot(1);zdot(1);ydotdot(1);zdotdot(1)];

%% Main Simulation

for i = 2:length(time)
    
    % Retrieve the command thrust from desired trajectory
    current_state = [y(i-1), z(i-1), theta(i-1), ydot(i-1), zdot(i-1), thetadot(i-1)];
    current_time = time(i);
    
    % Get our desired state at time(i)
    
    if traj_type == "cubic"
        if time(i) < end_time
            yz_temp = ppval(traj_obj,time(i));
            yzdot_temp = ppval(traj_obj_dot,time(i));
            yzdotdot_temp = ppval(traj_obj_dotdot,time(i));
        else
            yz_temp = waypoints(:,end);
            yzdot_temp = [0;0];
            yzdotdot_temp = [0;0];
        end
    elseif traj_type == "trim" || traj_type == "stepV" || traj_type == "stepA_FF"
        yzdotdot_temp = [0 ; 0];
        yzdot_temp = [V_s ; 0];
        yz_temp = [V_s*time(i-1) ; 0];
    elseif traj_type == "increasing"
        if time(i) < (end_time-buffer_time)
            yzdotdot_temp = [a_s ; 0];
            yzdot_temp = [a_s*time(i-1) ; 0];
            yz_temp = [(1/2)*a_s*(time(i-1)^2) ; 0];
        else
            yzdotdot_temp = [0 ; 0];
            yzdot_temp = [V_s ; 0];
            yz_temp = [(1/2)*a_s*((end_time-buffer_time)^2) + V_s*(time(i) - (end_time-buffer_time)) ; 0];
        end
    elseif traj_type == "decreasing"
        if time(i) < (end_time-buffer_time)
            yzdotdot_temp = [-a_s ; 0];
            yzdot_temp = [V_start-a_s*time(i-1) ; 0];
            yz_temp = [V_start*time(i-1)-(1/2)*a_s*(time(i-1)^2) ; 0];
        else
            yzdotdot_temp = [0 ; 0];
            yzdot_temp = [0 ; 0];
            yz_temp = [V_start*(end_time-buffer_time) - 0.5*a_s*(end_time-buffer_time)^2 ; 0];
        end
    elseif traj_type == "prescribed_aoa"
        % Take the trajectory generation section and read from there
        time_temp = round(end_time-buffer_time-dt,2);
        if time(i) < (end_time-buffer_time)
            yzdotdot_temp = [ydotdot_des(i); 0];
            yzdot_temp = [ydot_des(i); 0];
            yz_temp = [y_des(i); 0];
        else
            yzdotdot_temp = [0;0];
            yzdot_temp = [V_s ; 0];
            yz_temp = [y(time == time_temp) + V_s*(time(i) - (end_time-buffer_time)) ; 0];
        end
    elseif traj_type == "stepA_hover" || traj_type == "stepP"
        yzdotdot_temp = [0 ; 0];
        yzdot_temp = [0 ; 0];
        yz_temp = [0 ; 0];
    end
    
    desired_state(:,i) = [yz_temp' , yzdot_temp' , yzdotdot_temp']; % 6x1
    
    % Find the current airspeed and prop wash speed
    Vi(i-1) = sqrt( ydot(i-1)^2 + zdot(i-1)^2 );
    
    % Compute orientations
    if abs(Vi(i-1)) >= 1e-10
        gamma(i-1) = atan2(zdot(i-1), ydot(i-1));  % Inertial orientation
    else
        gamma(i-1) = 0;
    end
    alpha(i-1) = theta(i-1) - gamma(i-1);  % Angle of attack strictly based on inertial speed
    
    % Get prop wash over wing via momentum theory
    T_avg = 0.5*(T_top(i-1) + T_bot(i-1));
    
    %     Vw(i-1) = 1.2*sqrt( T_avg/(8*rho*pi*R^2) );
    Vw(i-1) = eta*sqrt( (Vi(i-1)*cos(theta(i-1)-gamma(i-1)))^2 + (T_avg/(0.5*rho*pi*R^2)) );
    Vw_top(i-1) = eta*sqrt( (Vi(i-1)*cos(theta(i-1)-gamma(i-1)))^2 + (T_top(i-1)/(0.5*rho*pi*R^2)) );
    Vw_bot(i-1) = eta*sqrt( (Vi(i-1)*cos(theta(i-1)-gamma(i-1)))^2 + (T_bot(i-1)/(0.5*rho*pi*R^2)) );
    
    % Compute true airspeed over the wings using law of cosines
    Va(i-1) = sqrt( Vi(i-1)^2 + Vw(i-1)^2 + 2*Vi(i-1)*Vw(i-1)*cos( alpha(i-1)) );
    
    % Use this check to avoid errors in asin
    if Va(i-1) >= 1e-10
        alpha_e(i-1) = asin(Vi(i-1)*sin(alpha(i-1))/Va(i-1));
    else
        alpha_e(i-1) = 0;
    end
    
    % Retrieve aero coefficients based on angle of attack
    if aero == true
        %         [Cl, Cd, Cm] = aero_fns(c0, c1, c2, alpha_e(i-1));
        %         Cl = interp1(alpha_data, cl_data, alpha_e(i-1)*180/pi);
        %         Cd = interp1(alpha_data, cd_data, alpha_e(i-1)*180/pi);
        %         Cm = interp1(alpha_data, cm_data, alpha_e(i-1)*180/pi);
        Cl = ppval(cl_spline, alpha_e(i-1)*180/pi);
        Cd = ppval(cd_spline, alpha_e(i-1)*180/pi);
        Cm = ppval(cm_spline, alpha_e(i-1)*180/pi);
    else
        Cl = 0;
        Cd = 0;
        Cm = 0;
        alpha_e(i-1) = alpha(i-1);  % The traditional angle of attack is now true.
    end
    
    % Compute aero forces/moments
    L(i-1) = 0.5*rho*Va(i-1)^2*(chord*span)*Cl;
    D(i-1) = 0.5*rho*Va(i-1)^2*(chord*span)*Cd;
    M_air(i-1) = 0.5*rho*Va(i-1)^2*(chord*span)*chord*Cm;
    
    
    % Controller
    [T_top(i), T_bot(i), Fdes(:,i)] = qbit_controller(current_state, ...
        desired_state(:,i), L(i-1), D(i-1), M_air(i-1), alpha_e(i-1), m, ...
        Ixx, l);
    
    if integrate_method == "euler"
        %%%%%%%%%% Euler Integration
        ydotdot(i) = ((T_top(i) + T_bot(i))*cos(theta(i-1)) - D(i-1)*cos(theta(i-1) - alpha_e(i-1)) - L(i-1)*sin(theta(i-1) - alpha_e(i-1)))/m;
        zdotdot(i) = ( -m*g + (T_top(i) + T_bot(i))*sin(theta(i-1)) - D(i-1)*sin(theta(i-1) - alpha_e(i-1)) + L(i-1)*cos(theta(i-1) - alpha_e(i-1)))/m;
        thetadotdot(i) = (M_air(i-1) + l*(T_bot(i) - T_top(i)))/Ixx;
        
        % Euler integration
        ydot(i) = ydot(i-1) + ydotdot(i)*dt;
        zdot(i) = zdot(i-1) + zdotdot(i)*dt;
        thetadot(i) = thetadot(i-1) + thetadotdot(i)*dt;
        
        y(i) = y(i-1) + ydot(i)*dt;
        z(i) = z(i-1) + zdot(i)*dt;
        theta(i) = theta(i-1) + thetadot(i)*dt;
        
    elseif integrate_method == "rk4"
        %%%%%%%%%%% 4th-Order Runge Kutta:
        state = [y(i-1);z(i-1);theta(i-1);ydot(i-1);zdot(i-1);thetadot(i-1)];
        k1 = dynamics(state, m, g, Ixx, l, T_top(i), T_bot(i), L(i-1), D(i-1), M_air(i-1), alpha_e(i-1));
        k2 = dynamics(state+(dt/2)*k1, m, g, Ixx, l, T_top(i), T_bot(i), L(i-1), D(i-1), M_air(i-1), alpha_e(i-1));
        k3 = dynamics(state+(dt/2)*k2, m, g, Ixx, l, T_top(i), T_bot(i), L(i-1), D(i-1), M_air(i-1), alpha_e(i-1));
        k4 = dynamics(state+(dt)*k3, m, g, Ixx, l, T_top(i), T_bot(i), L(i-1), D(i-1), M_air(i-1), alpha_e(i-1));
        
        new_state = state + (dt/6)*(k1 + 2*k2 + 2*k3 + k4);
        y(i) = new_state(1);
        z(i) = new_state(2);
        theta(i) = new_state(3);
        ydot(i) = new_state(4);
        zdot(i) = new_state(5);
        thetadot(i) = new_state(6);
        
        ydotdot(i) = k1(4);
        zdotdot(i) = k1(5);
        thetadotdot(i) = k1(6);
    else
        errordlg("Incorrect integration scheme")
    end
end

% Padding
L(end) = L(end-1);
D(end) = D(end-1);
M_air(end) = M_air(end-1);

Va(end) = Va(end-1);
Vi(end) = Vi(end-1);
Vw(end) = Vw(end-1);
Vw_top(end) = Vw_top(end-1);
Vw_bot(end) = Vw_bot(end-1);

alpha(end) = alpha(end-1);
alpha_e(end) = alpha_e(end-1);
gamma(end) = gamma(end-1);

T_top(end) = T_top(end-1);
T_bot(end) = T_bot(end-1);

Fdes(:,end) = Fdes(:,end-1);
Fdes(:,1) = Fdes(:,2);

alpha_e_startidx = find(alpha_e ~= 0,1,'first');
alpha_e(1:(alpha_e_startidx-1)) = alpha_e(alpha_e_startidx);

Va(1) = Va(2);
Vw(1) = Vw(2);
Vw_top(1) = Vw_top(2);
Vw_bot(1) = Vw_bot(2);
T_top(1) = T_top(2);
T_bot(1) = T_bot(2);
ydotdot(1) = ydotdot(2);
zdotdot(1) = zdotdot(2);

a_v_Va = (1/2)*rho*(chord*span)*Va.^2/(m*g);

%% Trim Comparison
% Take the data from the trim analysis for the particular flight condition
% we're interested in (based on eta)

table = readtable("prop_wash_sweep.csv");
trim_eta = table.eta;
trim_alpha_e = table.alpha_e(trim_eta == eta);
trim_theta = table.theta(trim_eta == eta);
trim_alpha = table.alpha(trim_eta == eta);
trim_Vi = table.V_i(trim_eta == eta);
trim_a_v_Va = table.a_v_Va(trim_eta == eta);
trim_Cl = table.Cl(trim_eta == eta);
trim_Cd = table.Cd(trim_eta == eta);

if traj_type == "increasing" || traj_type == "decreasing"
    % Apply the acceleration shift based on derivation of a_v relationship
    % with alpha.
    if traj_type == "decreasing"
        a_s = -a_s;
    end
    trim_a_v_Va_shift = trim_a_v_Va - (a_s/g)./(trim_Cd + trim_Cl.*cot(trim_alpha_e*pi/180));
end

%% Plotting
qbit_main_plotting()

%% Dynamics Function
function xdot = dynamics(x, m, g, Ixx, l, T_top, T_bot, L, D, M_air, alpha_e)
% INPUTS
% t - current time (time(i))
% x - current state , x = [6x1] = [y, z, theta, ydot, zdot, thetadot]
% m, g, Ixx, l - physical parameters of mass, gravity, inertia, prop arm
% length
% T_top, T_bot - motor thrust inputs
% L, D, M_air - aero forces and moments, computed prior
% alpha_e - effective AoA on the wing

xdot = zeros(size(x));

xdot(1) = x(4);
xdot(2) = x(5);
xdot(3) = x(6);
xdot(4) = ((T_top + T_bot)*cos(x(3)) - D*cos(x(3) - alpha_e) - L*sin(x(3) - alpha_e))/m;
xdot(5) = ( -m*g + (T_top + T_bot)*sin(x(3)) - D*sin(x(3) - alpha_e) + L*cos(x(3) - alpha_e))/m;
xdot(6) = (M_air + l*(T_bot - T_top))/Ixx;

end
\end{lstlisting}

\begin{lstlisting}[language=Octave,caption={This function houses the controller described in section \ref{sec:geometric_controller}.}, captionpos=t]
%%% This function will output thrust commands based on a nonlinear
%%% geometric controller that tracks orientation [theta] and position [x,z].
%%% Spencer Folk 2020
function [T_top, T_bot, Fdes] = qbit_controller(current_state, desired_state, L, D, M_air, alpha_e, m, Ixx, l)

% INPUTS -
% current_state = [x z theta xdot zdot thetadot]'
% current_time - current time step (time(i))
% L - current lift force
% D - current drag force
% M_air - current pitch moment
% alpha_e - current effective angle of attack
% m - vehicle mass
% Ixx - vehicle inertia about x axis
% l - distance between each rotor

%% Gains and constants
K_p = [5.8*2 , 0 ; 0 , 5.8*3];
% K_d = [8.41*2 , 0 ; 0 , 8.41*3];
K_d = 2*sqrt(K_p(1,1))*eye(2);

K_R = 373.6489/10;
% K_w = 19.333;
K_w = 2*sqrt(K_R);

g = 9.81;
max_motor_thrust = 0.30*9.81*2; % N, determined by estimating max thrust of a single motor and multiplying by 2

% Booleans -- for clarity, true nominally means it will be allowed or enabled.
aero = true;   % This bool determines whether or not the controller is aware of aerodynamic forces
neg_thrust_bool = true;  % Boolean for allowing negative thrusts by the motor (unrealistic, but for debugging purposes)
motor_sat_bool = false;  % If motor thrust goes above saturation limit, this will limit it.


%% Extract current and trajectory states for a given time
y = current_state(1);
z = current_state(2);
theta = current_state(3);
ydot = current_state(4);
zdot = current_state(5);
thetadot = current_state(6);

rT = desired_state;  % Put function here trajectory(current_time)
yT = rT(1);
zT = rT(2);
ydotT = rT(3);
zdotT = rT(4);
ydotdotT = rT(5);
zdotdotT = rT(6);

%% Construct rotation matrices
iRb = [cos(theta) , -sin(theta) ; sin(theta) , cos(theta)];
iRe = [cos(theta - alpha_e) , -sin(theta - alpha_e) ; sin(theta - alpha_e) , cos(theta - alpha_e)];

%% Computing u1

% Compute desired accelerations
rdotdot_des = [ydotdotT ; zdotdotT] - K_d*[ydot - ydotT ; zdot - zdotT] - K_p*[y - yT ; z - zT];

% Now compute Fdes
if aero == true
    Fdes = m*rdotdot_des + [0 ; m*g] - iRe*[-D ; L];
else
    Fdes = m*rdotdot_des + [0 ; m*g];
end

% From there compute u1 (magnitude)
b1 = iRb*[1;0];
u1 = b1'*Fdes;

%% Computing u2

% Solve for b1_des:
b1_des = Fdes/norm(Fdes);

% Compute error
% e_theta = acos(dot(b1,b1_des));
e_theta = -atan2(b1(1)*b1_des(2) - b1(2)*b1_des(1), b1(1)*b1_des(1) + b1(2)*b1_des(2));

% Compute u2:
u2 = Ixx*(-K_R*e_theta - K_w*thetadot) - M_air;

%% Computing actuator outputs
% We can readily solve for thrust by solving this system:
% [A]*[T_top ; T_bot] = [u1 ; u2]

T = inv([1 , 1 ; -l , l])*[u1 ; u2];

T_top = T(1);
T_bot = T(2);

% Negative thrust constraint
if neg_thrust_bool == false
    if T_top < 0
        T_top = 0;
    end
    if T_bot < 0
        T_bot = 0;
    end
end

% Motor saturation constraint
if motor_sat_bool == true
    if T_top >= max_motor_thrust
        T_top = max_motor_thrust;
    end
    if T_bot >= max_motor_thrust
        T_bot = max_motor_thrust;
    end
end


end
\end{lstlisting}

\begin{lstlisting}[language=Octave,caption={The role of this function is to generate a constant-height transition with a prescribed angle of attack versus time.}, captionpos=t]
%%% This function designs a planar trajectory (y(t), ydot(t), yddot(t))
%%% For a constant height transition maneuver, based on a given time-
%%% valued function of alpha_e.
%%% Spencer Folk 2020
function [y_des, ydot_des, ydotdot_des]=prescribed_aoa_traj_generator(dt,time,alpha_e_des,cl_spline, cd_spline,rho,m,g,chord,span,accel_bool)
% INPUTS
% dt - sampling rate
% time - time vector corresponding to alpha_e_des
% alpha_e_des - alpha_e (in rad) corresponding to the i'th simulation step
% rho - air density [kg/m^3]
% m - vehicle mass [kg]
% g - gravity [m/s^2]
% chord - wing chord [m]
% span - wing span [m]
% R - rotor radius [m]
% accel_bool - boolean to determine whether or not we consider the
% acceleration

y_des = zeros(1,length(alpha_e_des));
ydot_des = zeros(1,length(alpha_e_des));
ydotdot_des = zeros(1,length(alpha_e_des));

% Get Cl, Cd for the desired alpha_e
cl = ppval(cl_spline,alpha_e_des*180/pi);
cd = ppval(cd_spline,alpha_e_des*180/pi);


if accel_bool == 0
    %%%%%%%%%%%%%%%%% ANALYTICAL WITH NO ACCELERATION %%%%%%%%%%%%%%%%%%%%%%%%%
    % Compute V_i(t) from desired
    V_i = sqrt((2*m*g*cot(alpha_e_des)./(rho*chord*span*(cd + cl.*cot(alpha_e_des)))));
    
    % From V_i we can assign our values for the trajectory:
    ydot_des = V_i;
    ydotdot_des_temp = diff(V_i)/dt;
    ydotdot_des = [ydotdot_des_temp , ydotdot_des_temp(end)];
    
    for i = 2:length(alpha_e_des)
        y_des(i) = y_des(i-1) + V_i(i)*dt;
    end
    
elseif accel_bool == 1
    %%%%%%%%%%%%%%%%% APPROXIMATE SOLN WITH ACCELERATION %%%%%%%%%%%%%%%%%%%%%%
    V_i = zeros(size(time));
    
    for i = 2:length(time)
        V_idot = g*cot(alpha_e_des(i)) - ((0.5*rho*chord*span*(cl(i)*cos(alpha_e_des(i)) + cd(i)*sin(alpha_e_des(i))))/...
            (m*sin(alpha_e_des(i))))*V_i(i-1)^2;
        V_i(i) = V_i(i-1) + V_idot*dt;
        
        ydotdot_des(i) = V_idot;
        ydot_des(i) = V_i(i);
        y_des(i) = y_des(i-1) + V_i(i)*dt;
    end
end

end
\end{lstlisting}

\end{document}